\documentclass{article}

\usepackage{microtype}
\usepackage{graphicx}
\usepackage{subfigure}
\usepackage{booktabs} % for professional tables
\usepackage{enumitem}

\usepackage{hyperref}

\usepackage[accepted]{icml2025}

\usepackage{amsmath}
\usepackage{amssymb}
\usepackage{mathtools}
\usepackage{amsthm}
\usepackage{hyperref}
\usepackage[table,dvipsnames]{xcolor}
\newcommand{\blue}[1]{$_{\color{BlueGreen}\downarrow #1}$}
\newcommand{\red}[1]{$_{\color{RedOrange}\uparrow #1}$}
\usepackage{multirow}
\usepackage{pifont}

\newcommand{\cmark}{\ding{51}}
\newcommand{\xmark}{\ding{55}}

\usepackage{array}

\definecolor{rapsblue}{RGB}{0, 76, 153}
\definecolor{rapspurple}{RGB}{128, 0, 128}
\definecolor{rapsgreen}{RGB}{0, 100, 0}

\usepackage[capitalize,noabbrev]{cleveref}

\theoremstyle{plain}

\theoremstyle{definition}

\theoremstyle{remark}

\usepackage[textsize=tiny]{todonotes}

\icmltitlerunning{Towards Adaptive, Scalable, and Robust Coordination of LLM Agents}

\begin{document}

\twocolumn[
\icmltitle{Towards Adaptive, Scalable, and Robust Coordination of LLM Agents: \\ A Dynamic Ad-Hoc Networking Perspective}
% \icmltitle{Towards Adaptive, Scalable, and Robust Coordination of LLM Agents: \\ A Reputation-Aware Publish-Subscribe Paradigm}

% It is OKAY to include author information, even for blind
% submissions: the style file will automatically remove it for you
% unless you've provided the [accepted] option to the icml2025
% package.

% List of affiliations: The first argument should be a (short)
% identifier you will use later to specify author affiliations
% Academic affiliations should list Department, University, City, Region, Country
% Industry affiliations should list Company, City, Region, Country

% You can specify symbols, otherwise they are numbered in order.
% Ideally, you should not use this facility. Affiliations will be numbered
% in order of appearance and this is the preferred way.
\icmlsetsymbol{equal}{*}

\begin{icmlauthorlist}
\icmlauthor{Rui Li}{1}
\icmlauthor{Zeyu Zhang}{1}
\icmlauthor{Xiaohe Bo}{1}
\icmlauthor{Quanyu Dai}{2}
\icmlauthor{Chaozhuo Li}{3}
\icmlauthor{Feng Wen}{2}
\icmlauthor{Xu Chen}{1,equal}
%\icmlauthor{}{sch}
% \icmlauthor{Firstname8 Lastname8}{sch}
% \icmlauthor{Firstname8 Lastname8}{yyy,comp}
%\icmlauthor{}{sch}
%\icmlauthor{}{sch}
\end{icmlauthorlist}

\icmlaffiliation{1}{Gaoling School of Artificial Intelligence, Renmin University of China.}
\icmlaffiliation{2}{Huawei Technologies Ltd.}
% \icmlaffiliation{3}{Beijing University of Posts and Telecommunications}
\icmlaffiliation{3}{Beijing Academy of Artificial Intelligence}

\icmlcorrespondingauthor{Xu Chen}{}

% You may provide any keywords that you
% find helpful for describing your paper; these are used to populate
% the "keywords" metadata in the PDF but will not be shown in the document
\icmlkeywords{Machine Learning, ICML}

\vskip 0.3in
]

% this must go after the closing bracket ] following \twocolumn[ ...

% This command actually creates the footnote in the first column
% listing the affiliations and the copyright notice.
% The command takes one argument, which is text to display at the start of the footnote.
% The \icmlEqualContribution command is standard text for equal contribution.
% Remove it (just {}) if you do not need this facility.

\printAffiliationsAndNotice{}  % leave blank if no need to mention equal contribution
% \printAffiliationsAndNotice{\icmlEqualContribution} % otherwise use the standard text.

\begin{abstract}
Multi-agent architectures built on large language models (LLMs) have demonstrated the potential to realize swarm intelligence through well-crafted collaboration.
However, the substantial burden of manual orchestration inherently raises an imperative to automate the design of agentic workflows.
We frame such an agent coordination challenge as a classic problem in dynamic ad-hoc networking: \emph{How to establish adaptive and reliable communication among a scalable number of agentic hosts?}
In response to this unresolved dilemma, we introduce \textbf{RAPS}, a reputation-aware publish-subscribe paradigm for \emph{adaptive}, \emph{scalable}, and \emph{robust} coor\-dination of LLM agents.
RAPS is grounded in the \emph{Distributed Publish-Subscribe Protocol}, allowing LLM agents to exchange messages based on their declared intents rather than predefined topologies.
Beyond this substrate, RAPS further incorporates two coherent overlays: (i) \emph{Reactive Subscription}, enabling agents to dynamically refine their intents; and (ii) \emph{Bayesian Reputation}, empowering each agent with a local watchdog to detect and isolate malicious peers.
Extensive experiments over five benchmarks showcase that our design effectively reconciles adaptivity, scalability, and robustness in a unified multi-agent coordination framework.
\end{abstract}

\section{Introduction}

Multi-agent systems (MAS) \cite{park2023generative, li2023camel, chatdev, hong2023metagpt, tran2025multi_agent_survey} powered by Large Language Models (LLMs) \cite{llm1, llm4, dubey2024llama3} have emerged as a promising avenue towards swarm intelligence, allowing specialized LLM agents \cite{llm-agent-survey-2} to collaboratively tackle complex tasks beyond the reach of isolated models \cite{gpt-swarm}. Notably, the efficacy of these systems critically hinges on their underlying coordination architectures~\cite{qian2024scaling}.
% Early designs predominantly rely on manual orchestration with fixed interaction topologies, inherently incurring substantial engineering overhead and fails to generalize across shifting task distributions.
Early manual orchestration inherently incurs substantial engineering overhead and~struggles to generalize across shifting task distributions, thereby underscoring the imperative for \emph{automatic multi-agent coordination} \cite{gpt-swarm, g-designer, aflow, puppeteer}, which empowers the systems to autonomously orchestrate agentic workflows without human intervention.

\begin{figure}[t!]
\centering
\vspace{3mm}
\includegraphics[width=1.0\columnwidth]{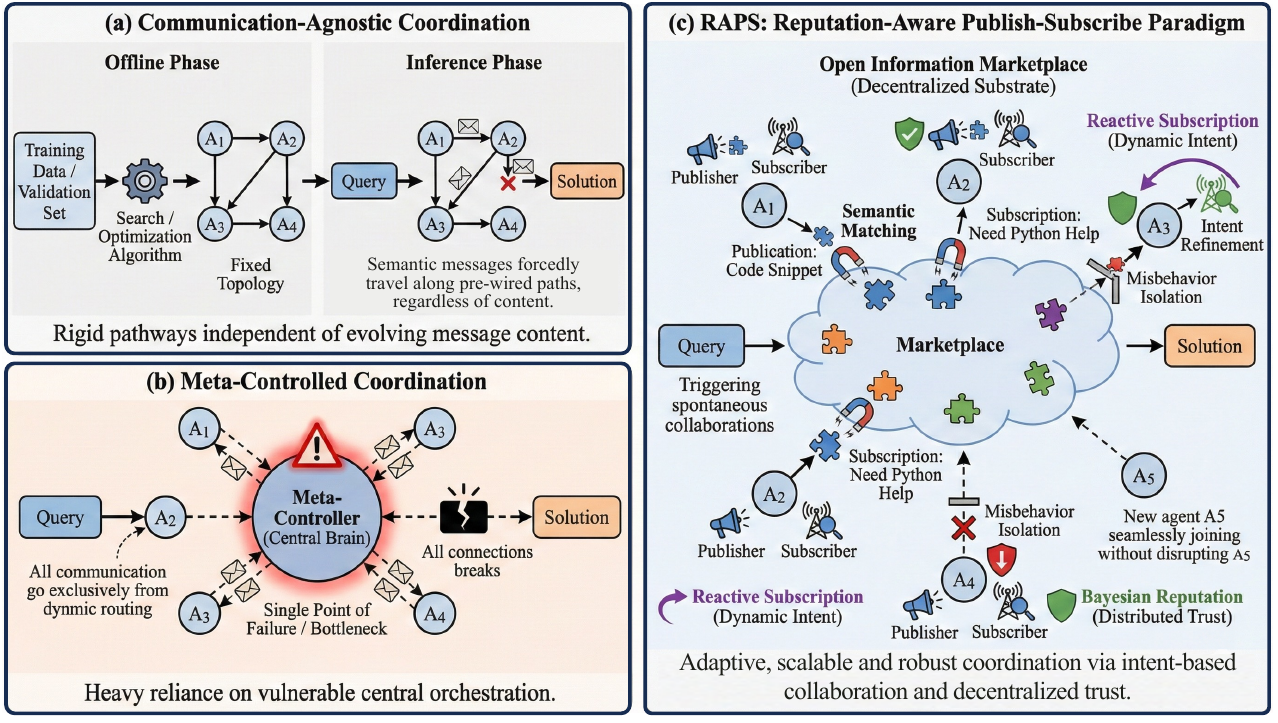}
\vspace{-7.5mm}
\caption{Illustrations of different MAS coordination paradigms.}
\label{Fig-1}
% \vskip -0.2in
\vspace{-5mm}
\end{figure}

In this paper, we approach the challenge of automatic coordination through the perspective of computer networking.
% we take a networking-inspired perspective on such an automatic coordination problem. 
Given their strong capabilities, LLM agents effectively function as network hosts that can perceive, process, and forward information.
Drawing this parallel, the coordination of LLM agents then naturally aligns with
% between MAS coordination and 
\emph{dynamic ad-hoc networking} \cite{ad_hoc-survey, ps-survey-1}: a set of agentic hosts actively communicate on demand without centralized control, while these hosts may join, leave, or misbehave.
Such an analogy directly informs our desiderata:
% From such a general standpoint, 
an ideal automatic MAS should \emph{(i) adapt to evolving message flow during execution, (ii) scale with dynamic membership discovery, and (iii) remain robust under misbehaving participants.} These three requirements---\emph{\textbf{adaptivity}}, \emph{\textbf{scalability}}, and \emph{\textbf{robustness}}---form a tightly coupled design space.

However, existing automatic coordination methods explore only parts of this design space, yet \emph{neither resolves the triad simultaneously}. 
Prior works shown in Figure 1(a) typically search for a specific agentic workflow \cite{gpt-swarm, adas, aflow} or train a parameterized coordination network \cite{g-designer, shen2025understanding} based on an offline validation set.
This paradigm imposes a fixed connectivity pattern independent of actual communication content (i.e., \emph{communication-agnostic}), rendering the system blind to the evolving message flow during inference.
% This paradigm decouples the coordination from communication for the sake of simplicity (i.e., \emph{communication-agnostic}), while failing to accommodate the evolving message flow during inference. 
Moreover, it lacks support for dynamic agent membership, forcing re-orchestration from scratch as agents join or leave.
% once the agent pool changes, whether agents join, leave, or misbehave, such a fixed architecture must be re-searched or re-trained.
% assumes static communication topologies that are blind to real-time message flows.
While several recent advances \cite{mas-zero, puppeteer, evoagent}, as depicted in Figure 1(b), leverage a high-level \emph{meta-controller} (e.g., a powerful LLM) to enable inference-time orchestration, their reliance on centralized control inherently incurs the single-point-of-failure bottleneck: If the high-authority meta-controller is compromised, the whole network may malfunction or even collapse, without effective mechanisms for recovery.
Therefore, a critical dilemma arises: \emph{How to reconcile adaptivity, scalability, and robustness within a unified coordination framework, without heavy training overhead or fragile central control?}

In light of this dilemma, we introduce \textbf{RAPS}, a \underline{R}eputation-\underline{A}ware \underline{P}ublish-\underline{S}ubscribe paradigm to establish the adaptive, scalable, and robust coordination among LLM-based agents, as shown in Figure 1(c).
Instead of enforcing a static topology, RAPS grounds its coordination fabric in a \emph{\textbf{Distributed Publish-Subscribe Protocol}} \cite{ps-survey-1, ps-2, ps-3}, which functionally decouples each agent host into three modules: a \emph{Subscriber} to state the agent's intents (i.e., \emph{subscriptions}); a \emph{Publisher} to execute the agentic function and generate new messages (i.e., \emph{publications}); and a \emph{Broker} to direct information flow by matching publications with content-aligned subscriptions.
% by dispatching publications to agents with subscriptions.
% based on the alignment between publications and subscriptions. 
This intent-based communication~protocol~serves as~an adaptive substrate that liberates the agents from rigid~interactions, fostering an open information marketplace where collaboration emerges spontaneously.
Building on the basic substrate, our RAPS framework further integrates two overlay mechanisms: (i) \emph{\textbf{Reactive Subscription}}, empowering the agents~to refine their subscriptions for \emph{dynamic intent evolution}; and 
(ii) \emph{\textbf{Bayesian Reputation}}, equipping each agent with a local watchdog to assess peer reliability via Bayesian estimation for \emph{trustworthy decentralized collaboration}.
By combining the publish-subscribe substrate and the overlay mechanisms, RAPS effectively reconciles the critical demands of adaptivity, scalability, and robustness within a unified framework.

Our contributions are summarized as follows:
\begin{itemize}[topsep=4pt, itemsep=0pt]
    \item \textbf{Perspective:} We frame the MAS coordination problem from a new perspective of dynamic ad-hoc networking, illuminating the key challenge of reconciling adaptivity, scalability, and robustness within a unified framework.
    \vspace{-10pt}
    \item \textbf{Communication Substrate:} We propose a \emph{Distributed Publish-Subscribe Protocol} as the coordination fabric. This protocol functionally decouples LLM agents into intent-declaring Subscribers and message-generating Publishers, liberating the system from rigid topologies to foster spontaneous content-centric collaboration.
    \item \textbf{Overlay Mechanisms:} We further design two critical mechanisms to augment the communication substrate: \emph{Reactive Subscription} and \emph{Bayesian Reputation}, where the former facilitates dynamic intent evolution and the latter enforces trustworthy decentralized collaboration.
    \item \textbf{Evaluation:} Extensive experiments across five diverse benchmarks demonstrate that our holistic RAPS framework effectively reconciles adaptivity, scalability, and robustness for multi-agent coordination, with consistent performance improvements over existing methods.
\end{itemize}

\vspace{-2mm}
\section{Background}

\textbf{From LLM Agents to MAS.}
The evolution of LLMs has catalyzed a paradigm shift from passive response generation to autonomous agents capable of perception, reasoning, and action \cite{react, toolformer, reflexion}. 
While single-agent models excel at isolated tasks, they typically struggle with complex problems that require diverse expertise or intricate planning \cite{gpt-swarm}. Inspired by the ``Society of Mind'' theory \cite{minsky1986society}, early frameworks such as CAMEL \cite{li2023camel}, MetaGPT \cite{hong2023metagpt}, and ChatDev \cite{chatdev} employ role-playing mechanisms or standard operating procedures to facilitate collaboration. However, these systems primarily rely on hard-coded rules and linear chains, lacking the flexibility to handle dynamic task requirements.

\begin{figure*}[t!]
\centering
\includegraphics[width=0.87\textwidth]{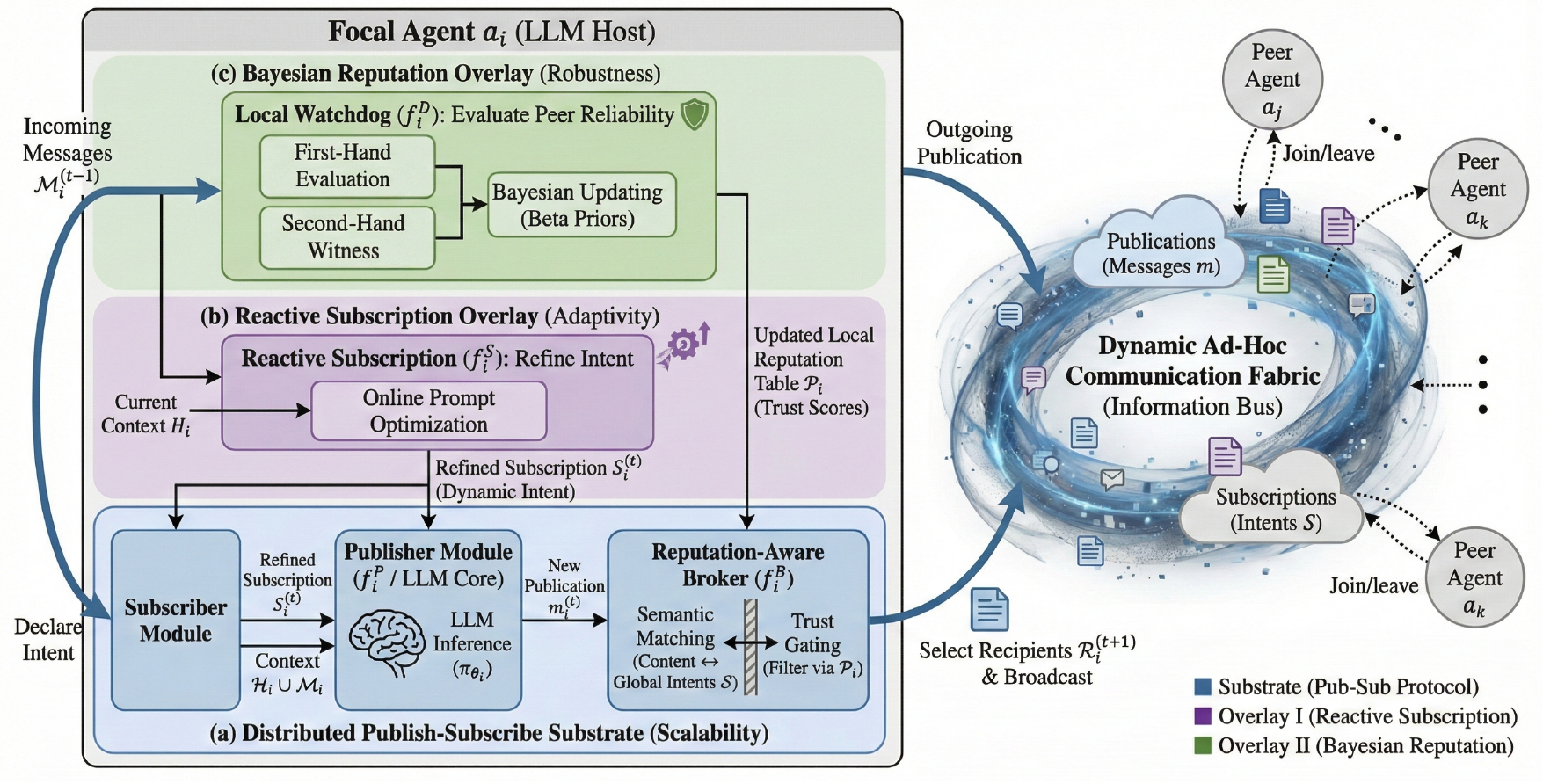}
\vspace{-2.5mm}
\caption{The overall framework of RAPS. The architecture consists of three layers: (a) A distributed publish-subscribe substrate that decouples agents into publishers and subscribers for MAS coordination; (b) Reactive subscription that allows agents to adaptively refine their intents based on the message flow; and (3) Bayesian reputation that employs a local watchdog to assess peer reliability for robustness.}
\label{Fig-2}
% \vskip -0.2in
\vspace{-3mm}
\end{figure*}

\textbf{Communication-Agnostic Automatic Coordination.}
To mitigate the burden of manual engineering, a growing body of research focuses on automating the coordination of LLM agents.
Pioneering works such as GPTSwarm \cite{gpt-swarm} model agent interactions as computation graphs, employing evolutionary algorithms to search for optimal connectivity patterns on a validation set.
Similarly, some frameworks like ADAS \cite{adas} and AFlow \cite{aflow} utilize search-based optimization to discover effective workflows, while G-Designer \cite{g-designer} and EIB-Learner \cite{shen2025understanding} train a graph neural network to model interaction patterns.
These methods decouple the topology orchestration from inference-time communication, inherently inducing rigid agentic workflows regardless of the specific inference context.
Such a \emph{communication-agnostic} strategy not only hinders the \textbf{\emph{adaptivity}} to evolving message flow, but also impedes the \textbf{\emph{scalability}} of candidate agents,~as agent membership changes would render the pre-wired topo\-logies obsolete, incurring substantial re-optimization costs.

\textbf{Meta-Controlled Automatic Coordination.} To address the rigidity of communication-agnostic coordination, a parallel stream of research adopts a centralized control paradigm to govern the agent communication.
Recent advances such~as AutoAgents \cite{chen2023autoagents}, MAS-Zero \cite{mas-zero}, and Puppeteer \cite{puppeteer} leverage a high-authority meta-planner to actively oversee the entire agent interactions and direct the message flow during inference.
While such~an inference-time orchestration strategy enables message-level adaptivity to some extent, its efficacy is tightly bound to the meta-controller's capabilities \cite{puppeteer},~incurring~a \textbf{\emph{robustness}} problem of single-point-of-failure.
% :~If~the~central controller hallucinates or is compromised, its derived coordination fabric inherently risks malfunction or collapse.~
Furthermore, this centralized dependency also imposes a severe bottleneck on \textbf{\emph{scalability}}, as the decision space for the meta-controller expands combinatorially as the agent population increases.

\textbf{In contrast to these methods,} RAPS shifts the coordination paradigm from offline topology search or fragile centralized control to a \emph{distributed intent-centric communication fabric}. By grounding LLM agent interactions in a publish-subscribe substrate enriched with two overlay mechanisms, our RAPS fosters spontaneous collaboration based on dynamic intents and peer trust, effectively reconciling adaptivity, scalability, and robustness within a unified framework.
Extended discus\-sions on broader related works are provided in Appendix~\ref{app:related_work}.

\section[LLM Agent Coordination as Dynamic Ad-Hoc Networking]
{\parbox[t]{0.9\columnwidth}{LLM Agent Coordination as \\ Dynamic Ad-Hoc Networking}}
\label{sec:3}

In this section, we formalize the multi-agent coordination problem and establish its correspondence with \emph{Dynamic Ad-Hoc Networking}. This perspective illuminates three critical desiderata (i.e., \emph{Adaptivity}, \emph{Scalability}, and \emph{Robustness}) that current coordination methods fail to satisfy simultaneously.

\subsection{Problem Formulation}

We formulate the multi-agent system as a dynamic population of agentic hosts $\mathcal{A} = \{a_1, \dots, a_N\}$. Each host $a_i\in\mathcal{A}$ operates under an LLM-driven policy $\pi_{\theta_i}$.
The operational context of $\pi_{\theta_i}$ is encapsulated by a configuration tuple $\mathcal{C}_i = \\\langle S_i, H_i, \mathcal{T}_i \rangle$, where $S_i$ denotes the system prompt specifying the role or function of $a_i$, 
$H_i$ represents the memory~buffer storing the accumulated knowledge and interaction history of $a_i$, and $\mathcal{T}_i$ is a collection of external tools available to $a_i$.
Given a complex task query $q$, the objective of the system is to synthesize a solution through a collaborative process.

At any discrete time step $t$, the agent $a_i$ observes the current state of its local context $H_i^{(t)}$ and then generates a message $m_i^{(t)} \sim \pi_{\theta_i}(\cdot | S_i, H_i^{(t)}, \mathcal{T}_i)$.
% $m_i^{(t)} \sim \pi_{\theta_i}(\cdot|H_i^{(t)})$.
The core challenge of MAS design centers on the \emph{coordination function} $\Phi$, which governs the dissemination scope of this new message:
% {\begingroup
% \fontsize{9.5pt}{11pt}\selectfont
\begin{equation}
\label{eq:1}
% \scalebox{0.96}{$\displaystyle
    \mathcal{R}_i^{(t+1)} = \Phi(m_i^{(t)}, \mathcal{A}), \; \text{where } \mathcal{R}_i^{(t+1)} \subseteq \mathcal{A} \setminus \{a_i\}.
% $}
\end{equation}
% \endgroup}
% Upon receiving the new message, 
When the new message arrives, each recipient $a_j \in \mathcal{R}_i^{(t+1)}$ updates its local context $H_j^{(t+1)} \leftarrow H_j^{(t)} \cup \{m_i^{(t)}\}$ and may subsequently trigger new actions. 
In contrast to recent works that typically model $\Phi$ as either a static topological mapping (i.e., routing messages on \emph{communication-agnostic graphs}) or a global orchestration policy (i.e., leveraging a~\emph{centralized meta-controller} to dispatch messages), we rethink the auto\-matic coordination problem from the perspective of \emph{dynamic ad-hoc networking}, where communication patterns emerge spontaneously based on the semantic intents of each agent.

\subsection{Networking Perspective}

We posit that the coordination of LLM agents is structurally isomorphic to the operation of \textbf{dynamic ad-hoc networks} \cite{ad_hoc-survey}. Specifically, we map the basic components of LLM-MAS to classic networking primitives:
\begin{itemize}[leftmargin=*]
    \vspace{-3.0mm}
    \item \textbf{Hosts ($a_i$):} Each LLM agent functions as an autonomous network host. Under this analogy, the system prompt~$\mathcal{S}_i$~of $a_i$ defines the host's service profile; the memory state $\mathcal{H}_i^{(t)}$ acts as the local cache for storing interaction history; and the toolset $\mathcal{T}_i$ represents the executable plugins for service.
    \vspace{-5.3mm}
    \item \textbf{Packets ($m_i^{(t)}$):} The natural language messages generated by agents serve as information packets carrying semantic payloads. Unlike traditional TCP/IP networks that route packets based on explicit destination addresses \cite{tcp_survey}, the dissemination of $m_i^{(t)}$ in dynamic ad-hoc networks is governed by the semantic alignment between its payload and the receivers' intents \cite{ccn_survey}.
    \vspace{-1.2mm}
    \item \textbf{Topology ($\mathcal{G}^{(t)}$):} The interaction graph $\mathcal{G}^{(t)}$~induced~by~$\Phi$ naturally reveals a transient communication pattern. In the ad-hoc settings, this topology is decoupled from physical infrastructure, operating instead as a logical overlay that evolves on-the-fly. Links between hosts emerge spontane\-ously, driven by the shifts in host intents \cite{ps-3} and peer trustworthiness \cite{confidant}.
    \vspace{-7.0mm}
\end{itemize}
This mapping also reveals that prevailing MAS orchestration works such as AFlow \cite{aflow} and G-Designer \cite{g-designer} are akin to the legacy \emph{circuit-switching} networks \cite{network_book}, which are incompatible with the dynamic information flow of swarm intelligence. To~deal with this rigidity, the system necessitates a paradigm shift towards \emph{content-centric packet-switching} \cite{eugster2003many}, where the routing decisions are decentralized and driven by semantic content rather than pre-established paths.

\textbf{Design Desiderata.} 
Drawing on the isomorphism between MAS and dynamic ad-hoc networking, we posit an effective coordination design to be one that reconciles three critical properties:
(i) \emph{Adaptivity} requires intent-based communica\-tion, where message dissemination is governed~by~semantic alignment rather than fixed topologies that are agnostic to the evolving message flows~\cite{gpt-swarm, maas, g-designer, aflow};
(ii) \emph{Scalability} demands the support for dynamic membership~\cite{eugster2003many} to accommodate variable agent populations, avoiding the prohibitive re-optimization overhead of pre-wiring models \cite{gpt-swarm, maas, g-designer, aflow} and the combinatorial bottlenecks of centralized meta-controllers \cite{mas-zero, puppeteer};  
(iii) \emph{Robustness} necessitates a decentralized trust mechanism to insulate the system from adversarial behaviors \cite{adv_survey} and alleviate the single-point-of-failure risks \cite{mas-zero}. 
Refer to Appendix \ref{app:related_work} for more discussions.

\section{RAPS: Reputation-Aware Publish-Subscribe Communication}
\label{sec:method}

In this paper, we introduce \textbf{RAPS}, a new MAS coordination framework that simultaneously reconciles the desiderata of adaptivity, scalability, and robustness outlined in Section~\ref{sec:3}.
We first provide an overview of our RAPS framework (\S \ref{sec:overview}).
Subsequently, we elaborate on its communication substrate (\S \ref{sec:substrate}) and two critical overlay mechanisms (\S \ref{sec:overlay I} and \S \ref{sec:overlay II}).

\subsection{Overview}
\label{sec:overview}

As illustrated in Figure~\ref{Fig-2}, RAPS fundamentally shifts the coordination paradigm from rigid topology orchestration or fragile centralized control to intent-centric communication.

At its core, RAPS employs a \emph{distributed publish-subscribe protocol} as the communication substrate, which functionally decouples the agentic hosts into packet producers and con\-sumers, directing the information flow based on the semantic alignment between declared intents (i.e., \emph{subscriptions}) and produced contents (i.e., \emph{publications}). 
By eschewing~the~pre-wired interaction graphs and the vulnerable central planners, this substrate fosters an open information marketplace for adaptive MAS coordination, while also allowing the agents to seamlessly join or leave without topology reconfiguration.

Building on such a communication substrate, RAPS further incorporates two overlay mechanisms to facilitate dynamic intent evolution and trustworthy decentralized collaboration. 
\emph{Reactive Subscription} empowers the agents to dynamically update their intents based on the evolving message contexts, akin to the online prompt optimization \cite{li2025prompt_survey}~that can spontaneously adapt to the unfolding interaction history.
\emph{Bayesian Reputation} equips each agent with a local watch\-dog that maintains probabilistic beliefs about peer reliability to insulates the system from adversarial behaviors, ensuring robust collaboration without recourse to a central authority.
By integrating these overlays atop the distributed substrate, RAPS coherently fulfills the desiderata of adaptivity, scala\-bility, and robustness in a unified coordination framework.

\subsection{\textsc{Substrate}: Publish-Subscribe Protocol}
\label{sec:substrate}

To dismantle the rigidity of pre-wired interaction topologies, RAPS moves beyond current MAS designs by establishing~a distributed publish-subscribe protocol as its communication substrate.
% Under this protocol, 
This protocol functionally decouples the agentic hosts into message producers (\emph{Publishers}) and consumers (\emph{Subscribers}), mediated by a content-aware \emph{Broker} module.

\textbf{Three Functional Roles.} 
Formally, we define the operational tuple for each agent $a_i \in \mathcal{A}$ as $\langle f_i^\mathtt{S}, f_i^\mathtt{P}, f_i^\mathtt{B} \rangle$, corresponding to three distinct \emph{logical roles}:
\begin{itemize}[leftmargin=*, topsep=0pt, itemsep=0pt]
    \item \textbf{Subscriber ($f_i^\mathtt{S}$):} This function allows the network host to explicitly declare its packet intents or service capability. In the agentic context, such a subscription naturally aligns with the system prompt $S_i$ configured when each agent $a_i$ joins the network. For instance, a host initialized with the role ``\texttt{Python Expert}'' directly broadcasts a standing subscription to the network, signaling its ability to process the publications about code generation or debugging tasks.
    \item \textbf{Publisher ($f_i^\mathtt{P}$):} The publisher acts as the functional core of agent $a_i$. 
    At step $t$, upon receiving a packet $m^{(t-1)}$~that aligns with its subscription $S_i$, the publisher directly runs the base policy $\pi_{\theta_i}$ conditioned on its local context buffer $\langle S_i, H_i^{(t-1)} \cup m^{(t-1)}, \mathcal{T}_i \rangle$ to produce a new message $m^{(t)}$. Note that the publisher does not specify recipients of $m^{(t)}$, but rather broadcasts it as a semantic signal for alignment.
    \item \textbf{Broker ($f_i^\mathtt{B}$):} This module operates as the instantiation of the coordination function $\Phi$~in~Equation~(\ref{eq:1}).
    Unlike prior methods that model $\Phi$ as static lookup or central planning, the broker implements $\Phi$ through \emph{decentralized semantic matching}: it derives the recipient set $\mathcal{R}_i^{(t+1)}$ by assessing the compatibility between the published message $m_t$ and the global subscription pool $\mathcal{S} = \{S_j\}_{j \neq i}$.
    Such an intent-content bridging process fosters the interaction graph $\mathcal{G}^{(t)}$ to emerge spontaneously in response to the task progress.
\end{itemize}

\textbf{Coordination Dynamics.}
With these three functional roles, the basic coordination protocol of RAPS alternates between \emph{Publication} and \emph{Brokerage} to induce the interaction patterns.

\emph{Publication Phase.} At any time step $t$, RAPS operates on a set of active agents $\mathcal{R}_*^{(t)}$. 
For the initial step ($t=0$), $\mathcal{R}_{\text{init}}^{(0)}$ is populated with default entry roles (e.g., an \texttt{Analyst} agent) activated directly by the input query $q$, whereas $\mathcal{R}_*^{(t)}$ for the subsequent steps is inherited from the preceding brokerage outcomes.
Every active agent $a_i \in \mathcal{R}_*^{(t)}$ first aggregates the received message packets $\mathcal{M}_i^{(t-1)}=\{m_*^{(t-1)} \mid a_i \in \mathcal{R}_*^{(t)}\}$ to update its interaction context. 
Then, $a_i$ runs the publisher function $f_i^\mathtt{P}$ to produce the new publication:
\begin{equation}
\label{eq_pub}
    % m_i^{(t)} \sim \pi_{\theta_i}(\cdot \mid S_i, \underbrace{H_i^{(t-1)} \cup \mathcal{M}_i^{(t-1)}}_{\text{Updated Context}},\mathcal{T}_i).
    m_i^{(t)} = f_i^\mathtt{P}\!\left( S_i,~ H_i^{(t-1)} \cup \mathcal{M}_i^{(t-1)}\right).
\end{equation}
In practice, $f_i^\mathtt{P}$ is typically instantiated as the LLM backbone $\pi_{\theta_i}$ of $a_i$. The publication $m_i^{(t)}$ serves as a flexible semantic container modulated by the output instructions of policy $\pi_{\theta_i}$, capable of explicitly encapsulating the chain-of-thought reasoning \cite{wei2022chain}, intermediate execution results,~and strategic plans intended for subsequent interactions.

\emph{Brokerage Phase.} Upon the publication of $m_i^{(t)}$, the control flow shifts to the broker $f_i^\mathtt{B}$ to determine dissemination path for this message.
The broker treats $m_i^{(t)}$ as a query to search for compatible consumers within the standing subscriptions $\mathcal{S}$ of other hosts.
Formally, for each producer $a_i$, its broker identifies a subset of recipients $\mathcal{R}_{i}^{(t+1)}$:
\begin{equation}
\label{eq_bro}
    \mathcal{R}_{i}^{(t+1)} = f_i^\mathtt{B}\!\left(m_i^{(t)}, \mathcal{S}\right) \subseteq \mathcal{A}.
\end{equation}
In practice, the instantiation of $f_i^\mathtt{B}$ spans a spectrum of complexity, ranging from lightweight embedding-based scoring to expressive LLM-driven selection.
Such brokers can only determine one recipient each time to produce dynamic chain topology \cite{puppeteer} for simple scenarios, while also retaining the flexibility to activate multiple subscribers sim\-ultaneously for complex problem solving \cite{li2025knowtrace}.

\emph{Termination.} This publication-brokerage process terminates when the active agents emit a consensus finish signal or the maximum time step limit is reached.
Finally, the system agg\-regates all publications and delegating one terminal agent~to provide the final answer \cite{jiang2023llm-blender, wu2024autogen, dylan, zhang2024cut, g-designer}.

This intent-driven substrate naturally establishes an adaptive and scalable communication fabric by liberating message flow from static topology.
Notably, the raw protocol neither accounts for the evolving granularity of agent subscriptions nor guards against potential misbehavior in an open system.
To bridge these gaps, we further augment this substrate with two coherent overlay mechanisms: \emph{reactive subscription} for dynamic intent evolution (\S \ref{sec:overlay I}) and \emph{Bayesian reputation} for trustworthy decentralized collaboration (\S \ref{sec:overlay II}).

\subsection{\textsc{Overlay I}: Reactive Subscription}
\label{sec:overlay I}

Under the substrate protocol, LLM agents act as static hosts with fixed subscriptions (i.e., system prompts) initialized at their joining time. However, such seed subscriptions~may~be too abstract or even absent for the emerging message types.

To improve the intent-content alignment without introducing centralized control, RAPS augments the basic substrate with a lightweight mechanism of \emph{Reactive Subscription}, allowing each active agent to refine its standing subscription based on the newly received packets before the publication phase.

\textbf{Reactive Subscription.} At time step $t$, each active host~$a_i\in\mathcal{R}_*^{(t)}$ first updates its system profile $S_i^{(t-1)}$ according to the latest messages $\mathcal{M}_i^{(t-1)}$ and optionally its context $H_i^{(t-1)}$:
\begin{equation}
    S_i^{(t)}=f^\mathtt{S}_i\!\left(S_i^{(t-1)},\, H_i^{(t-1)} \cup \mathcal{M}_i^{(t-1)}\right),
\end{equation}
where $f_i^\mathtt{S}$ is instantiated as an LLM-driven prompt rewriter that specializes the seed subscription by identifying salient intents, constraints, or tool cues from the received messages.
Such an updated $S_i^{(t)}$ then directly conditions the subsequent publication in Equation (\ref{eq_pub}), replacing the static prompt with the reactive one, so that the next message is generated under a more context-aware role to improve the functional quality.
Moreover, RAPS can optionally promote the transient local profile $S_i^{(t)}$ into the global subscription pool $\mathcal{S}$ under certain confirmation signals (e.g., peer acknowledgments in the~next subsection), thus coherently realizing \emph{dynamic membership discovery} without any centralized topology reconfiguration.

\subsection{\textsc{Overlay II}: Bayesian Reputation}
\label{sec:overlay II}

While the reactive subscription mechanism improves intent-content alignment, the open-ended communication substrate of RAPS remains vulnerable to \emph{unreliable behaviors} such~as hallucinations \cite{hall_survey}, tool misuse \cite{fu2024imprompter}, or adversarial manipulation \cite{adv_survey}. 
To secure the coordination without reverting to a centralized global judge, RAPS presents a fully decentralized \emph{Bayesian Reputation Overlay} inspired by classic robustness designs \cite{buchegger2002performance_confidant, confidant} in ad-hoc networks.

\textbf{Distributed Watchdog.}
RAPS equips each agent $a_i$ with~a \emph{local watchdog} that maintains pairwise probabilistic beliefs about every peer $a_j$ it interacts with. 
Concretely, the watchdog of $a_i$ holds three probabilistic ratings for $a_j$, including \emph{First-Hand Rating} $F_{ij}$ that indicates $a_i$'s direct evaluation of $a_j$'s behavioral quality; \emph{Trust Rating} $T_{ij}$ that quantifies~how much $a_i$ trusts $a_j$ as a reporter of reputation evidences; and the overall reputation $P_{ij}$ that aggregates direct evaluation and indirect reports into the final belief about $a_j$'s reliability.

\textbf{Bayesian Modeling.} 
Since LLM agents operate with sparse, noisy, and non-stationary context, the point estimates of peer reliability can be brittle.
% these probabilistic ratings are inherently uncertainty-aware.
% To draw calibrated 
To draw uncertainty-aware~beliefs,
% that can be updated online and naturally encode confidence, 
we cast each rating as a Bayesian posterior over an underlying Bernoulli parameter associated with the evidence~stream.
% From the perspective of $a_i$, 
% each rating for peer $a_j$ corresponds to a Bernoulli variable with its own latent parameter.
Following the classic settings \cite{davison2003statistical_1, berger2013statistical_2}, we then use the conjugate Beta-Bernoulli model to represent the uncertainty of each rating with two pseudo-counts:
\begin{equation}
\scalebox{0.93}{$\displaystyle
F_{ij}=\beta(x^F_{ij},y^F_{ij}),\ 
T_{ij}=\beta(x^T_{ij},y^T_{ij}),\ 
P_{ij}=\beta(x^P_{ij},y^P_{ij}).
$}
\end{equation}
Here $F_{ij}$ and $P_{ij}$ summarize $a_j$'s behavioral reliability as a publication source, while $T_{ij}$ represents the credibility of $a_j$, each driven by its corresponding Bernoulli evidence stream.

All ratings start from a non-informative prior $\beta(1,1)$, and are updated incrementally as the new observation $s_{ij}^{*}$ arrives.
Since agent behaviors and task regimes may drift over time, 
we adopt a discounted update that emphasizes the recent $s_{ij}^*$:\!
% observation $s_{ij}^*\in\{0,1\}$:
\begin{equation}
\label{eq_rule}
x_{ij}^* \leftarrow \lambda\,x_{ij}^* + s_{ij}^*, \qquad
y_{ij}^* \leftarrow \lambda\,y_{ij}^* + (1-s_{ij}^*),
\end{equation}
where $s_{ij}^*$ denotes the corresponding binary evidence for the specific rating to be updated (i.e., $*\in\{F,T,R\}$); $\lambda\in(0,1]$ represents a decay factor applied to the past pseudo-counts. 
The posterior mean $\mathbb{E}[\beta(x^*_{ij},y^*_{ij})]=x^*_{ij}/(x^*_{ij}+y^*_{ij})$ offers a calibrated point estimate, while the concentration $(x^*_{ij}+y^*_{ij})$ also reflects epistemic uncertainty for downstream~decisions.\!

\textbf{Reputation Calculation.}
Based on the Bayesian modeling, RAPS calculates each rating by instantiating the Bernoulli evidence stream during coordination.
When $a_i$ routes a message to a downstream agent $a_j$, the local watchdog of $a_i$ (i) performs a \emph{first-hand evaluation} on the publication of $a_j$, (ii) selectively collects \emph{second-hand} testimonials from other witnesses and assesses their credibility to update trust, and (iii) merges admissible direct/indirect evidence into the final reputation posterior that fuels the reputation-aware brokers.

\emph{First-Hand Evaluation.}
Whenever $a_i$ receives a publication from $a_j$, the local watchdog of~$a_i$ (e.g., an LLM auditor) performs an on-the-fly verification to produce a first-hand binary evaluation score $s^{F}_{ij}\in\{0,1\}$.
Concretely, $s^{F}_{ij}=0$ if the publication of $a_j$ is judged useful and consistent with its current context; otherwise $s^{F}_{ij}=1$ if the watchdog detects qualified misbehaviors such as factual error and tool misuse:
\begin{equation}
s^{F}_{ij}=f_i^\mathtt{D}\!\left(m_j^{(t)},\, S_i^{(t)},\, H_i^{(t)}\right)\in\{0,1\},
\end{equation}
where $f_i^\mathtt{D}$ denotes the watchdog function of $a_i$.
The evidence $s^{F}_{ij}$ is then used to update the first-hand posterior $F_{ij}=\beta(x^F_{ij},y^F_{ij})$ via the discounted rule in Equation (\ref{eq_rule}), yielding a direct rationality summary of $a_j$ from the perspective of~$a_i$.\!

\emph{Second-Hand Witness.}
While first-hand evaluation anchors reputation to verifiable local evidence, relying solely on $s^{F}_{ij}$ can be sample-inefficient in open coordination \cite{michiardi2002core}.
To accelerate the discovery of unreliable peers, RAPS allows $a_i$ to optionally solicit \emph{second-hand} testimonies from other witnesses that have recently interacted with $a_j$.
Concretely, upon receiving $m_j^{(t)}$, $a_i$ queries a small set of witnesses $\mathcal{W}_{ij}^{(t)}\subseteq\mathcal{A}\setminus\{a_i,a_j\}$ and obtains their first-hand posteriors about $a_j$, i.e.,
$\{F_{kj}=\beta(x^F_{kj},y^F_{kj})\}_{a_k\in\mathcal{W}_{ij}^{(t)}}$.
% instead of their already-merged reputation scores, so as to mitigate rumor amplification.
Since second-hand reports may be noisy or malicious, $a_i$ evaluates each witness $a_k$ by performing a \emph{deviation test} \cite{confidant} on its current belief $P_{ij}$:
\begin{equation}
\label{eq:deviation}
\frac{\left| \mathbb{E}[F_{kj}] - \mathbb{E}[P_{ij}] \right|}{\sqrt{\text{Var}(F_{kj}) + \text{Var}(P_{ij})}} \ge \delta,
\end{equation}
where $\delta$ is a predefined tolerance threshold. If the deviation exceeds $\delta$, the testimony is deemed incompatible (yielding evidence $s^{T}_{ik}=1$); otherwise, it is accepted ($s^{T}_{ik}=0$). This binary evidence is then used to update the trust posterior $T_{ik}$ via Eq.~(\ref{eq_rule}), progressively isolating dishonest witnesses.

\emph{Reputation Merging.}
To synthesize a holistic view of peer reliability, $a_i$ aggregates its direct evidence $s^{F}_{ij}$ with indirect testimonies into the cumulative reputation posterior $P_{ij}$ via \emph{linear pool model merging} \cite{berger2013statistical_2}.
First, $P_{ij}$ is anchored to the first-hand evidence $s^{F}_{ij}$, directly updating the parameters $(x^P_{ij},y^P_{ij})$ with Eq.~(\ref{eq_rule}).
Then, $a_i$ incorporates second-hand reports from a subset of \emph{admissible witnesses} $\mathcal{W}^*_{ij} \subseteq \mathcal{W}_{ij}^{(t)}$.
Following the classic protocol \cite{confidant}, a witness $a_k$ is deemed admissible if it is either locally trusted (i.e., $\mathbb{E}[T_{ik}] \ge \tau$) or its testimony successfully passes the deviation test (i.e., $s^{T}_{ik}=0$).
For every $a_k \in \mathcal{W}^*_{ij}$, we assimilate its reported posterior $F_{kj}=\beta(x^F_{kj},y^F_{kj})$ as weak pseudo-counts to refine the local belief:
\begin{equation}
x^{P}_{ij} \leftarrow x^{P}_{ij} + \omega\,x^{F}_{kj}, \qquad
y^{P}_{ij} \leftarrow y^{P}_{ij} + \omega\,y^{F}_{kj},
\end{equation}
where $\omega \in (0, 1)$ is a discounting factor.
Such a merging mechanism ensures that indirect evidence can modulate but does not overwhelm first-hand observations, resulting in a trust-gated aggregation that is robust to reputation poisoning.\!

\begin{table*}[t]
\caption{Performance comparisons with four types of baselines, including single-agent models, static multi-agent models, communication-agnostic models, and meta-controlled models. The best results are in \textbf{bold}, and the second best results are \underline{underlined}. All methods, except for the single-agent models, employ \textbf{five} identically configured agents for fair comparisons.}
\label{tab:main_results}
\vspace{1mm}
\centering
\small
\setlength{\tabcolsep}{5pt}
\setlength{\aboverulesep}{0pt} 
\setlength{\belowrulesep}{0pt} 
\renewcommand{\arraystretch}{1.1}
\resizebox{0.78\textwidth}{!}{
\begin{tabular}{l@{\hspace{5mm}}ccccc|c}
\toprule
\rule[-0.9ex]{0pt}{3.5ex}\textbf{Method} & \textbf{MMLU} & \textbf{GSM8K} & \textbf{SVAMP} & \textbf{AQuA} & \textbf{HumanEval} & \textbf{Average} \\
\midrule
\rowcolor{gray!15}\multicolumn{7}{l}{\textit{Single-Agent Models}} \\
% \midrule
Vanilla IO & 81.7 & 91.6 & 87.5 & 71.3 & 72.8 & 81.0 \\
CoT \cite{wei2022chain} & 83.0\red{1.3} & 92.1\red{0.5} & 88.4\red{0.9} & 74.7\red{3.4} & 75.7\red{2.9} & 82.8\red{1.8} \\
ComplexCoT \cite{complexcot} & 83.7\red{2.0} & 92.5\red{0.9} & 89.2\red{1.7} & 76.1\red{4.8} & 75.2\red{2.4} & 83.3\red{2.3} \\
SC \cite{wang2022self-consistency} & 82.4\red{0.7} & 92.4\red{0.8} & 88.9\red{1.4} & 76.8\red{5.5} & 77.5\red{4.7} & 83.6\red{2.6} \\
\midrule
\rowcolor{gray!15}\multicolumn{7}{l}{\textit{Static Multi-Agent Models}} \\
% \midrule
Chain \cite{qian2024scaling} & 84.3\red{2.6} & 91.7\red{0.1} & 82.6\blue{4.9} & 70.4\blue{0.9} & 81.3\red{8.5} & 82.1\red{1.1} \\
Star \cite{qian2024scaling} & 80.4\blue{1.3} & 91.9\red{0.3} & 88.2\red{0.7} & 69.6\blue{1.7} & 74.5\red{1.7} & 80.9\blue{0.1} \\
Tree \cite{qian2024scaling} & 82.4\red{0.7} & 90.7\blue{0.9} & 88.5\red{1.0} & 73.9\red{2.6} & 72.4\blue{0.4} & 81.6\red{0.6} \\
Random \cite{qian2024scaling} & 85.6\red{3.9} & 92.0\red{0.4} & 87.0\blue{0.5} & 75.1\red{3.8} & 78.2\red{5.4} & 83.6\red{2.6} \\
LLM-Debate \cite{llm-debate} & 85.0\red{3.3} & 92.4\red{0.8} & 89.8\red{2.3} & 77.3\red{6.0} & 82.6\red{9.8} & 85.4\red{4.4} \\
LLM-Blender \cite{jiang2023llm-blender} & 81.0\blue{0.7} & 91.3\blue{0.3} & 88.3\red{0.8} & 76.9\red{5.6} & -- & 84.4\red{3.4} \\
\midrule
\rowcolor{gray!15}\multicolumn{7}{l}{\textit{Communication-Agnostic Models}} \\
% \midrule
GPTSwarm \cite{gpt-swarm} & 83.7\red{2.0} & 92.7\red{1.1} & 88.5\red{1.0} & 78.2\red{6.9} & 88.5\red{15.7} & 86.3\red{5.3} \\
AgentPrune \cite{zhang2024cut} & 84.3\red{2.6} & 92.3\red{0.7} & 89.8\red{2.3} & 79.1\red{7.8} & 86.8\red{14.0} & 86.5\red{5.5} \\
AFlow \cite{aflow} & 85.6\red{3.9} & \underline{94.1}\red{2.5} & 90.0\red{2.5} & 78.5\red{7.2} & \underline{91.0}\red{18.2} & 87.8\red{6.8} \\
MaAS \cite{maas} & 85.0\red{3.3} & 91.4\blue{0.2} & 89.3\red{1.8} & 76.2\red{4.9} & 87.1\red{14.3} & 85.8\red{4.8} \\
G-Designer \cite{g-designer} & \underline{86.3}\red{4.6} & 93.2\red{1.6} & \underline{90.7}\red{3.2} & \underline{79.4}\red{8.1} & 90.2\red{17.4} & \underline{88.0}\red{7.0} \\
\midrule
\rowcolor{gray!15}\multicolumn{7}{l}{\textit{Meta-Controlled Models}} \\
% \midrule
AutoAgents \cite{chen2023autoagents} & 82.4\red{0.7} & 92.5\red{0.9} & 86.7\blue{0.8} & 75.7\red{4.4} & 87.6\red{14.8} & 85.0\red{4.0} \\
Puppeteer \cite{puppeteer} & 84.3\red{2.6} & 93.3\red{1.7} & 89.5\red{2.0} & 77.5\red{6.2} & 75.3\red{2.5} & 84.0\red{3.0} \\
MAS-Zero \cite{mas-zero} & 83.0\red{1.3} & 92.6\red{1.0} & 87.3\blue{0.2} & 72.9\red{1.6} & 83.9\red{11.1} & 83.9\red{2.9} \\
\midrule
\rowcolor{gray!15}\multicolumn{7}{l}{\textit{Dynamic Ad-Hoc Networking Perspective}} \\
\rule[-0.5ex]{0pt}{3.0ex}\textbf{RAPS (Ours)} & \textbf{88.2\red{6.5}} & \textbf{95.4\red{3.8}} & \textbf{92.2\red{4.7}} & \textbf{82.6\red{11.3}} & \textbf{91.5\red{18.7}} & \textbf{90.0\red{9.0}} \\
\bottomrule
\end{tabular}}
\vspace{-2mm}
\end{table*}

\textbf{Reputation-Aware Brokerage.}
With the reputation overlay, the broker of $a_i$ evolves to be reliability-aware, thus enabling us to rewrite the brokerage function in Equation (\ref{eq_bro}) as:
\begin{equation}
\mathcal{R}_{i}^{(t+1)} = f_i^\mathtt{B}\!\left(m_i^{(t)},\, \mathcal{S},\, \mathcal{P}_i\right)\subseteq\mathcal{A},
\end{equation}
where $\mathcal{P}_i$ denotes $a_i$'s local reputation table.
As a result, the information flow is steered away from untrusted hosts, preventing local failures from cascading into global corruption.

\begin{figure}[t] % [t] 表示优先放置在当前栏的顶部
\vspace{-2mm}
    \centering    \includegraphics[width=\columnwidth]{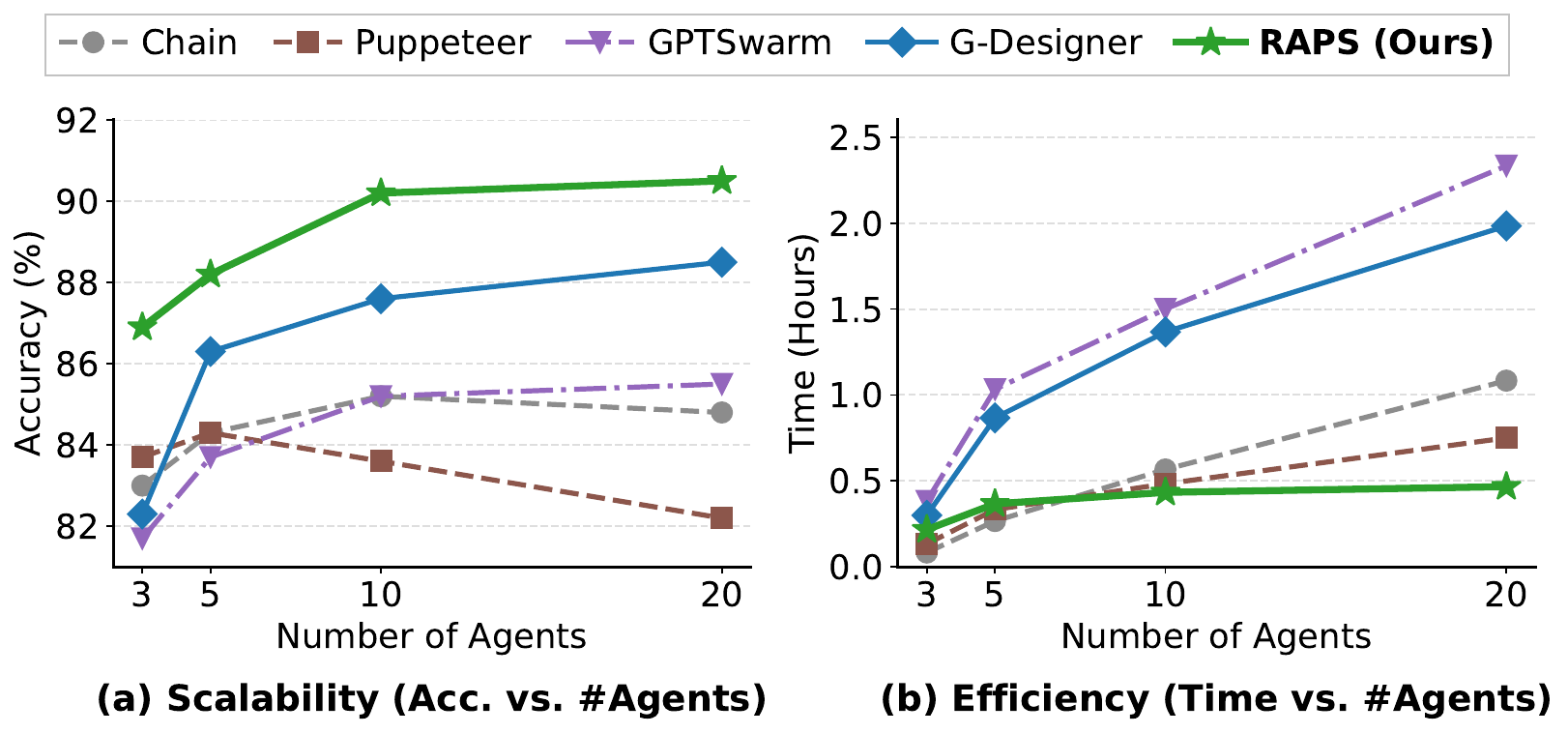}
    \vspace{-8mm}
    \caption{Scalability and Efficiency analysis with varying \#\hspace{0.3mm}agents.}
    \label{fig:scalability}
    \vspace{-3mm} 
\end{figure}

% \vspace{-3mm}
\section{Experiments}
\label{sec:exp}

To comprehensively validate the effectiveness of our RAPS framework, we conduct extensive experiments on a range of representative agent tasks spanning general reasoning,~math\-ematical problem solving, and code generation.
In the main experiment (\S\ref{sec:exp_main}), we compare RAPS with strong baselines over five heterogeneous benchmarks to assess its adaptivity.
Then, we vary the number of agents to study scalability and efficiency under open-ended memberships (\S\ref{sec:exp_scale}).
Next, we stress-test robustness (\S\ref{sec:exp_robust}) by injecting adversarial agents.
Finally, the detailed mechanism analysis is provided in \S\ref{sec:exp_ablation}.

\begin{figure}[t] % [t] 表示优先放置在当前栏的顶部
\vspace{-2mm}
    \centering    \includegraphics[width=\columnwidth]{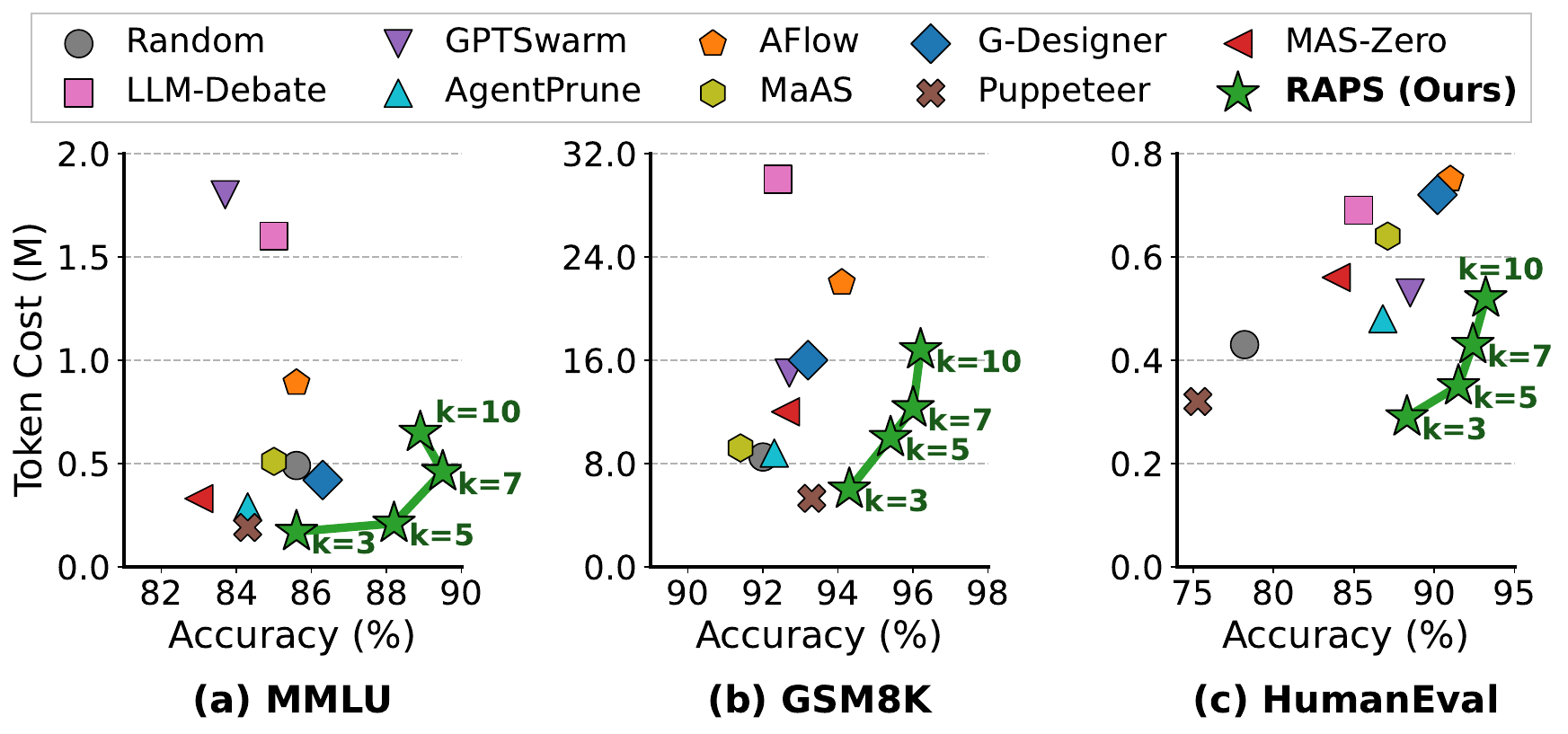}
    \vspace{-8.05mm}
    \caption{Cost-Performance analysis of different MAS coordination methods. $k$ represents the communication rounds for RAPS.}
    \label{fig:pareto}
    \vspace{-5mm} 
\end{figure}

\subsection{Experimental Setup}
\label{sec:exp_setup}

\textbf{Benchmarks.}
We evaluate RAPS over five standard datasets from three categories:
MMLU \cite{mmlu} for general reasoning;
GSM8K \cite{gsm8k}, SVAMP \cite{svamp}, and AQuA \cite{aqua} for mathematical reasoning;
HumanEval \cite{humaneval} for code generation.
We report Pass@1 for HumanEval and accuracy (Acc.) for others.
See Appendix~\ref{app:dataset} for the detailed statistics.

\textbf{Baselines.}
We compare RAPS with four types of baselines, including (1) \emph{Single-Agent Models}: CoT \cite{wei2022chain}, ComplexCoT \cite{complexcot}, and Self-Consistency (abbr. SC) \cite{wang2022self-consistency}; (2) \emph{Static Multi-Agent Models}: MacNet (Chain, Star, Tree, Random) \cite{qian2024scaling}, LLM-Debate \cite{llm-debate}, and LLM-Blender \cite{jiang2023llm-blender}; (3) \emph{Communication-Agnostic Coordination}: GPTSwarm \cite{gpt-swarm}, AgentPrune \cite{zhang2024cut}, AFlow \cite{aflow}, MaAS \cite{maas}, and G-Designer \cite{g-designer}; (4) \emph{Meta-Controlled Coordination}: AutoAgents \cite{chen2023autoagents}, Puppeteer \cite{puppeteer}, MAS-Zero \cite{mas-zero}.

\textbf{Implementation Details.}
Unless otherwise specified, all methods share the same closed-source commercial LLM as the backbone to ensure fair comparisons. For the agentic profiles in MAS, we adopt the standard role definitions and system prompt configurations following previous works~\cite{dylan, gpt-swarm}.
For our RAPS framework, we instantiate the publishers $f_i^\mathtt{P}$, reactive subscribers $f_i^\mathtt{S}$, and local watchdogs $f_i^\mathtt{D}$ using the same LLM backbone.
The broker $f_i^\mathtt{B}$ is implemented to support both embedding-based matching with a lightweight commercial text embedding model for efficiency and LLM-based reasoning for complex decisions.
More implementation details are provided in Appendix~\ref{app:details}.

\begin{table}[t]
\vspace{-2mm}
\caption{Robustness to \emph{After-Training} prompt attacks on MMLU. ``T'' and ``A'' represent truthful and adversarial agents, respectively.}
\label{tab:prompt_attack}
\vspace{1mm}
\centering
\small
\resizebox{0.87\columnwidth}{!}{
\begin{tabular}{lccccc}
\toprule
\textbf{Method} & \textbf{5T0A} & \textbf{4T1A} & \textbf{3T2A} & \textbf{2T3A} & \textbf{5T5A} \\
\midrule
Chain & 84.3 & 72.5 & 50.3 & 22.2 & 16.3 \\
Random & 85.6 & 81.7 & 35.3 & 18.3 & 45.1 \\
LLM-Debate & 85.0 & 78.4 & 62.1 & 30.7 & 47.7 \\
GPTSwarm & 83.7 & 75.2 & 55.6 & 23.5 & 52.9 \\
AFlow & 85.6 & 79.7 & 52.3 & 19.6 & 28.8 \\
G-Designer & 86.3 & 80.4 & 37.9 & 15.0 & 49.7 \\
Puppeteer-P & 84.3 & 77.8 & 65.4 & 32.0 & 51.6 \\
Puppeteer-C & 84.3 & 13.7 & - & - & - \\
\midrule
\emph{RAPS w/o BR} & 86.9 & 83.7 & 69.3 & 33.3 & 53.6 \\
\textbf{RAPS} & \textbf{88.2} & \textbf{87.6} & \textbf{84.3} & \textbf{83.0} & \textbf{86.3} \\
\bottomrule
\end{tabular}}
\vspace{-5mm}
\end{table}

\subsection{Main Results}
\label{sec:exp_main}

The performance comparison on five benchmarks is reported in Table~\ref{tab:main_results}.
RAPS achieves state-of-the-art performance across all tasks, attaining the highest average score of \textbf{90.0\%}. 
% This represents a substantial improvement over the strongest single-agent baseline (+6.4\% vs. SC) and the previous state-of-the-art method (+2.0\% vs. G-Designer).
We attribute these advanced gains to the message-level adaptivity of RAPS compared to existing paradigms:

\textbf{Superiority over Static Structures.}
The static models (e.g., Chain, Star) often lag behind even single-agent baselines (e.g., Chain is 4.9\% lower than CoT on SVAMP), indicating that rigid topologies cannot accommodate the diverse reasoning patterns required by different queries.
RAPS overcomes this limitation by allowing the interaction graph to emerge dynamically. The intent-based publish-subscribe substrate ensures that messages are routed to relevant agents, avoiding the structural mismatch inherent in static graphs.

\textbf{Advantages over Offline Optimization.}
Communication-agnostic methods such as AFlow and G-Designer search for or generate workflows without considering specific commu\-nication process.
% optimize workflows based on training sets but freeze the topology during inference.
% While effective, these methods lack flexibility for instance-specific variations.
Despite their effectiveness, such methods remain blind to the evolving message flow during inference.
Beyond these baselines, RAPS framework achieves superior performance (e.g., +3.2\% over G-Designer on AQuA) with its inherent capacity for inference-time coordination.
% RAPS outperforms these methods (e.g., +3.2\% over G-Designer on AQuA) because it performs \emph{inference-time} coordination. 
More\-over, the agents in RAPS can further tailor their roles to~the message context via \emph{reactive subscription}, thereby capturing subtle reasoning paths overlooked by the pre-wired models.

% by refining intents on-the-fly via the reactive subscription mechanism, agents in RAPS can tailor their roles to the dynamic task context, capturing subtle reasoning steps that fixed workflows might miss.

\textbf{Comparison with Centralized Control.}
Meta-controlled architectures (e.g., AutoAgents, Puppeteer, and MAS-Zero) rely on a central planner, inherently inducing a performance bottleneck due to the complexity of global state supervision.
RAPS surpasses these models (e.g., +5.0\% over AutoAgents on average) by distributing the routing logic.
Departing from a monolithic agent that dictates holistic connectivity, RAPS decentralizes the routing logic by empowering local brokers to adaptively align publications with relevant subscriptions, leading to more nuanced and effective collaboration patterns.\!

\begin{table}[t]
\vspace{-2mm}
\caption{Ablation study on different variants of RAPS.}
\label{tab:ablation}
\centering
\small
\vspace{1mm}
\resizebox{0.9\columnwidth}{!}{
\begin{tabular}{l|ccc}
\toprule
\textbf{Variant} & \textbf{MMLU} & \textbf{GSM8K} & \textbf{HumanEval} \\
\midrule
\textbf{RAPS (Full)} & \textbf{88.2} & \textbf{95.4} & \textbf{91.5} \\
\midrule
w/o RS & 85.6 ($\downarrow$2.6) & 93.7 ($\downarrow$1.7) & 89.3 ($\downarrow$2.2) \\
w/o BR & 86.9 ($\downarrow$1.3) & 94.5 ($\downarrow$0.9) & 90.7 ($\downarrow$0.8) \\
w/o Both & 83.7 ($\downarrow$4.5) & 92.8 ($\downarrow$2.6) & 88.5 ($\downarrow$3.0) \\
\midrule
w/ LLM Broker & 89.5 & 95.9 & 93.7 \\
w/ Naive Agent Pool & 85.0 & 93.5 & 90.2 \\
\bottomrule
\end{tabular}}
\vspace{-6mm}
\end{table}

\vspace{-1mm}
\subsection{Scalability and Efficiency}
\label{sec:exp_scale}

A critical requirement for dynamic ad-hoc networking is the ability to scale effectively as the number of hosts increases. We evaluate the scalability and efficiency of RAPS by varying the agent population size $N$ on the MMLU benchmark.

\textbf{Population Scalability.}
Figure~\ref{fig:scalability}(a) shows the performance (accuracy) of each method as the agent population increases.
The chain model suffers from severe accuracy degradation due to error propagation across increasing interaction hops.
Similarly, the Puppeteer model also fails to maintain perfor\-mance, since its central orchestrator struggles to analyze~and coordinate among the growing number of agents. In contrast, RAPS consistently outperforms such approaches,~exhibiting steady accuracy gains with the increasing number of agents. This demonstrates the scalability of our~distributed publish-subscribe protocol, confirming its ability to handle growing agent populations without compromising performance.

\textbf{Communication Efficiency.}
Figure~\ref{fig:scalability}(b) further reports the end-to-end runtime as the agent population scales. 
The two communication-agnostic approaches (i.e., GPTSwarm and G-Designer) inherently incur rapidly growing optimization overhead (over 2 hours for 20 agents) as the agent candidate pool expands.
In comparison, RAPS is \emph{training-free} and~per\-forms coordination entirely at inference time via lightweight publish-subscribe matching.
Moreover, the \emph{broker}-selective dissemination effectively prevents unnecessary interactions, 
thereby keeping the inference latency stable as $N$ increases. 
As a result, RAPS achieves favorable runtime scaling while simultaneously improving accuracy, demonstrating an efficient coordination substrate for open-membership systems.

\textbf{Cost-Performance Trade-off.}
Furthermore, we analyze the economic viability of our RAPS in Figure~\ref{fig:pareto}, which plots the Pareto frontier of accuracy versus token consumption on three representative benchmarks.
The green curve presents the performance of RAPS at varying communication rounds $k \in \{3, 5, 7, 10\}$.
One can observe that RAPS establishes a superior frontier compared to the state-of-the-art baselines,
highlighting its advantages in both accuracy and efficiency.

\begin{figure}[t]
    \centering
\includegraphics[width=0.9\columnwidth]{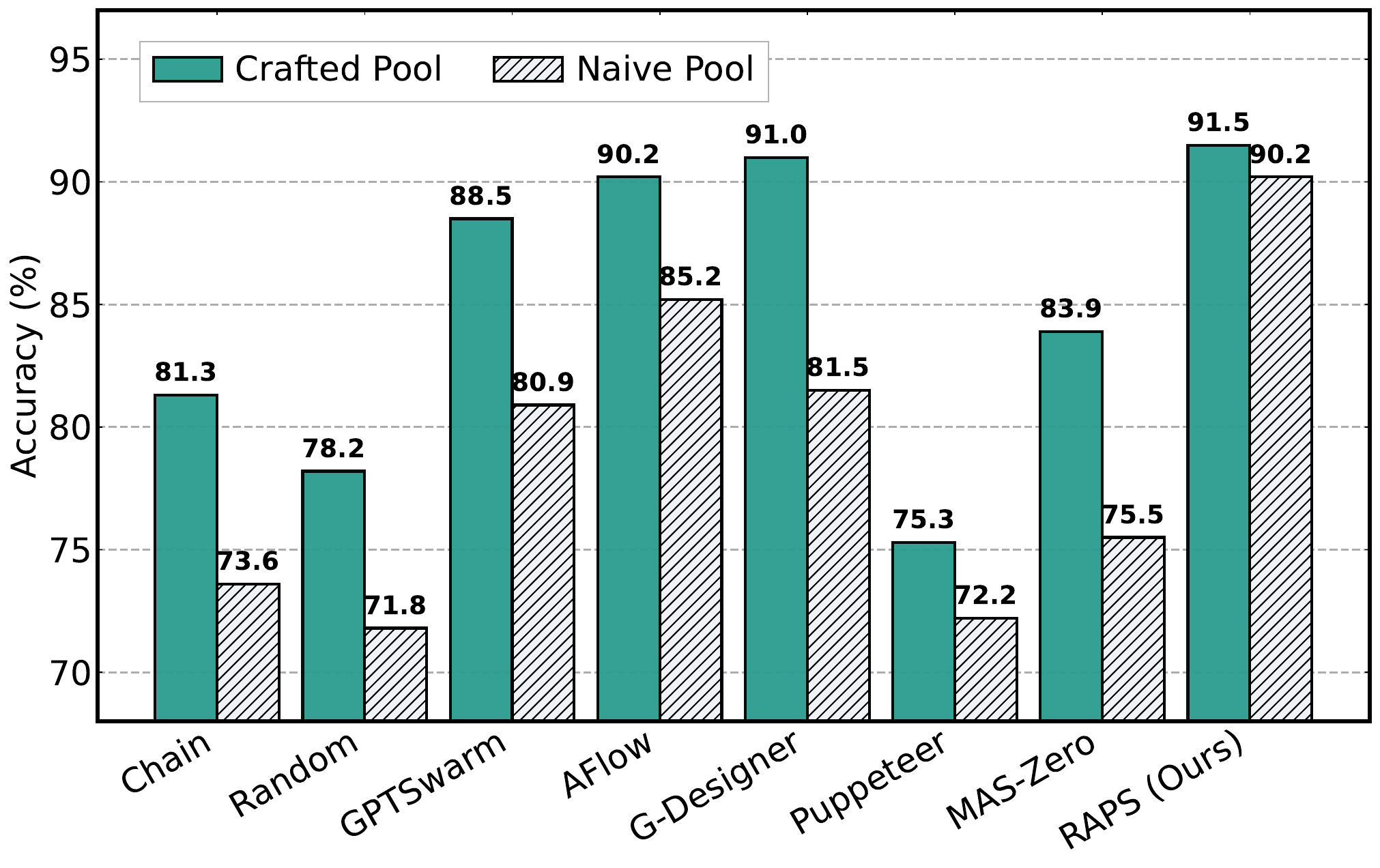}
    \vspace{-5mm}
    \caption{Impact of agent pool quality on HumanEval benchmark.}
    \label{fig:pool_impact}
    \vspace{-8mm}
\end{figure}

\subsection{Robustness Analysis}
\label{sec:exp_robust}

In open multi-agent systems, agents may exhibit unreliable behaviors due to hallucinations or adversarial attacks. To evaluate robustness, we conduct a ``Byzantine'' stress test on the MMLU benchmark by injecting adversarial agents prompted to provide misleading reasoning.
We vary the composition of the agent pool from clean (5 Truthful, 0 Adversarial) to highly contaminated (2 Truthful, 3 Adversarial, or a mix of 5T5A). The results are summarized in Table~\ref{tab:prompt_attack}.

\textbf{Fragility of Baselines.}
Existing models exhibit severe vulnerability to adversarial interference.
% As the proportion of adversarial agents increases, performance drops significantly across all baseline models.
For instance, AFlow and GPTSwarm experience sharp drops in accuracy when adversarial agents are introduced, with accuracy plummeting to as low as 19.6\% under the 2T3A scenario.
Moreover, Puppeteer, which relies on a centralized meta-controller, is susceptible to adversarial attacks in two different configurations: one where the adversarial prompt targets the non-central agents, and another where the adversarial prompt is aimed at the central meta-controller itself. In the latter case, we observe that Puppeteer-C collapses immediately (84.3\% $\to$ 13.7\%) with just one adversarial agent (4T1A), confirming that entrusting a monolithic high-authority controller is inherently unsafe for open coordination.

\textbf{Efficacy of Bayesian Reputation.}
In contrast, RAPS consistently outperforms all other methods, maintaining high accuracy even as the number of adversarial agents increases.
The performance of RAPS with and without Bayesian Reputation (BR) shows that the inclusion of BR can significantly enhance communication robustness, ensuring stable performance even in the contaminated agent pools.
These results demonstrate the effectiveness of BR in maintaining system reliability under adversarial conditions.

\subsection{Ablation Analysis}
\label{sec:exp_ablation}

To understand the contribution of each component within RAPS, we conduct an ablation study on MMLU, GSM8K, and HumanEval, as presented in Table~\ref{tab:ablation} and Figure~\ref{fig:pool_impact}.

\textbf{Impact of Overlay Mechanisms.}
We evaluate the impact of removing the two designed overlay mechanisms: Reactive Subscription (RS) and Bayesian Reputation (BR). As shown in Table~\ref{tab:ablation}, removing RS leads to a notable performance drop, highlighting the importance of dynamic subscriptions in adapting to the task dynamics; similarly, removing BR also incurs a moderate decline, as the reputation-aware brokerage can enhance the communication quality by filtering out low-quality reasoning signals. Therefore, these two mechanisms contribute synergistically to the full performance of RAPS.

\textbf{Broker Instantiation.}
We compare the default embedding-based broker with a more expressive LLM-based broker (\emph{w/ LLM Broker}). While the LLM broker yields marginal performance improvements (e.g., +1.3\% on MMLU), it would inherently incur higher communication latency. This justifies our design choice of using the embedding-based broker for the main experiments to balance efficiency and~accuracy.

\textbf{Resilience to Agent Pool Quality.}
A common limitation of current models is their sensitivity to the quality of initial agent configurations.
We investigate this by comparing the model performance on a \emph{Crafted Pool} (expert roles) versus a \emph{Naive Pool} (generic roles). As shown in Figure~\ref{fig:pool_impact},
existing models suffer severe degradation when using the naive pool.
For instance, G-Designer drops by 9.5\% (91.0\% $\to$ 81.5\%).
In contrast, \textbf{RAPS} exhibits remarkable resilience, maintain\-ing robust performance with only a negligible 1.3\% drop (91.5\% $\to$ 90.2\%).
This resilience is directly attributable to the \emph{Reactive Subscription} mechanism, which allows even naively initialized agents to autonomously refine their intents and specialize their roles based on the message context.

\section{Conclusion}

This paper re-envisions the coordination of LLM agents through the lens of dynamic ad-hoc networking, identifying the critical challenge of simultaneously achieving adaptivity, scalability, and robustness.
To address this dilemma, we introduced \textbf{RAPS}, a decentralized framework grounded in a reputation-aware publish-subscribe protocol.
By decoupling coordination from rigid topologies and fragile centralized controllers, RAPS empowers agents to spontaneously align their intents with evolving tasks and autonomously isolate unreliable peers via a distributed trust mechanism.
Extensive experiments across five benchmarks demonstrate that RAPS effectively reconciles the coordination triad, yielding state-of-the-art performance while scaling efficiently to larger agent populations and exhibiting superior resilience against adversarial interference.
We believe this networking-centric perspective provides a robust foundation for building open, self-organizing, and trustworthy multi-agent societies.

\section*{Impact Statement}

This paper presents work whose goal is to advance the field of 
Machine Learning. There are many potential societal consequences 
of our work, none which we feel must be specifically highlighted here.

% The above statement can be used verbatim in such cases, but we 
% encourage authors to think about whether there is content which does 
% warrant further discussion, as this statement will be apparent if the 
% paper is later flagged for ethics review.

% In the unusual situation where you want a paper to appear in the
% references without citing it in the main text, use \nocite
\nocite{langley00}

\bibliography{example_paper}

@inproceedings{langley00,
 author    = {P. Langley},
 title     = {Crafting Papers on Machine Learning},
 year      = {2000},
 pages     = {1207--1216},
 editor    = {Pat Langley},
 booktitle     = {Proceedings of the 17th International Conference
              on Machine Learning (ICML 2000)},
 address   = {Stanford, CA},
 publisher = {Morgan Kaufmann}
}

@inproceedings{park2023generative,
  title={Generative agents: Interactive simulacra of human behavior},
  author={Park, Joon Sung and O'Brien, Joseph and Cai, Carrie Jun and Morris, Meredith Ringel and Liang, Percy and Bernstein, Michael S},
  booktitle={Proceedings of the 36th annual acm symposium on user interface software and technology},
  pages={1--22},
  year={2023}
}

@inproceedings{hong2023metagpt,
title={Meta{GPT}: Meta Programming for A Multi-Agent Collaborative Framework},
author={Sirui Hong and Mingchen Zhuge and Jonathan Chen and Xiawu Zheng and Yuheng Cheng and Jinlin Wang and Ceyao Zhang and Zili Wang and Steven Ka Shing Yau and Zijuan Lin and Liyang Zhou and Chenyu Ran and Lingfeng Xiao and Chenglin Wu and J{\"u}rgen Schmidhuber},
booktitle={The Twelfth International Conference on Learning Representations},
year={2024}
}

@article{chatdev,
  title={Communicative agents for software development},
  author={Qian, Chen and Cong, Xin and Yang, Cheng and Chen, Weize and Su, Yusheng and Xu, Juyuan and Liu, Zhiyuan and Sun, Maosong},
  journal={arXiv preprint arXiv:2307.07924},
  volume={6},
  number={3},
  pages={1},
  year={2023}
}

@inproceedings{aflow,
title={{AF}low: Automating Agentic Workflow Generation},
author={Jiayi Zhang and Jinyu Xiang and Zhaoyang Yu and Fengwei Teng and Xiong-Hui Chen and Jiaqi Chen and Mingchen Zhuge and Xin Cheng and Sirui Hong and Jinlin Wang and Bingnan Zheng and Bang Liu and Yuyu Luo and Chenglin Wu},
booktitle={The Thirteenth International Conference on Learning Representations},
year={2025}
}

@inproceedings{gpt-swarm,
  title={Gptswarm: Language agents as optimizable graphs},
  author={Zhuge, Mingchen and Wang, Wenyi and Kirsch, Louis and Faccio, Francesco and Khizbullin, Dmitrii and Schmidhuber, J{\"u}rgen},
  booktitle={Forty-first International Conference on Machine Learning},
  year={2024}
}

@inproceedings{g-designer,
title={G-Designer: Architecting Multi-agent Communication Topologies via Graph Neural Networks},
author={Guibin Zhang and Yanwei Yue and Xiangguo Sun and Guancheng Wan and Miao Yu and Junfeng Fang and Kun Wang and Tianlong Chen and Dawei Cheng},
booktitle={Forty-second International Conference on Machine Learning},
year={2025}
}

@inproceedings{maas,
title={Multi-agent Architecture Search via Agentic Supernet},
author={Guibin Zhang and Luyang Niu and Junfeng Fang and Kun Wang and LEI BAI and Xiang Wang},
booktitle={Forty-second International Conference on Machine Learning},
year={2025}
}

@inproceedings{puppeteer,
title={Multi-Agent Collaboration via Evolving Orchestration},
author={Yufan Dang and Chen Qian and Xueheng Luo and Jingru Fan and Zihao Xie and Ruijie Shi and Weize Chen and Cheng Yang and Xiaoyin Che and Ye Tian and Xuantang Xiong and Lei Han and Zhiyuan Liu and Maosong Sun},
booktitle={The Thirty-ninth Annual Conference on Neural Information Processing Systems},
year={2025}
}

@inproceedings{evoagent,
  title={Evoagent: Towards automatic multi-agent generation via evolutionary algorithms},
  author={Yuan, Siyu and Song, Kaitao and Chen, Jiangjie and Tan, Xu and Li, Dongsheng and Yang, Deqing},
  booktitle={Proceedings of the 2025 Conference of the Nations of the Americas Chapter of the Association for Computational Linguistics: Human Language Technologies (Volume 1: Long Papers)},
  pages={6192--6217},
  year={2025}
}

@inproceedings{li2025knowtrace,
  title={Knowtrace: Bootstrapping iterative retrieval-augmented generation with structured knowledge tracing},
  author={Li, Rui and Dai, Quanyu and Zhang, Zeyu and Chen, Xu and Dong, Zhenhua and Wen, Ji-Rong},
  booktitle={Proceedings of the 31st ACM SIGKDD Conference on Knowledge Discovery and Data Mining V. 2},
  pages={1470--1480},
  year={2025}
}

@inproceedings{li2023camel,
  title={Camel: Communicative agents for" mind" exploration of large language model society},
  author={Li, Guohao and Hammoud, Hasan and Itani, Hani and Khizbullin, Dmitrii and Ghanem, Bernard},
  booktitle={Advances in Neural Information Processing Systems},
  volume={36},
  pages={51991--52008},
  year={2023}
}

@inproceedings{llm1,
 author = {Brown, Tom and Mann, Benjamin and Ryder, Nick and Subbiah, Melanie and Kaplan, Jared D and Dhariwal, Prafulla and Neelakantan, Arvind and Shyam, Pranav and Sastry, Girish and Askell, Amanda and Agarwal, Sandhini and Herbert-Voss, Ariel and Krueger, Gretchen and Henighan, Tom and Child, Rewon and Ramesh, Aditya and Ziegler, Daniel and Wu, Jeffrey and Winter, Clemens and Hesse, Chris and Chen, Mark and Sigler, Eric and Litwin, Mateusz and Gray, Scott and Chess, Benjamin and Clark, Jack and Berner, Christopher and McCandlish, Sam and Radford, Alec and Sutskever, Ilya and Amodei, Dario},
 booktitle = {Advances in Neural Information Processing Systems},
 pages = {1877--1901},
 title = {Language Models are Few-Shot Learners},
 volume = {33},
 year = {2020}
}

@article{dubey2024llama3,
  title={The llama 3 herd of models},
  author={Dubey, Abhimanyu and Jauhri, Abhinav and Pandey, Abhinav and Kadian, Abhishek and Al-Dahle, Ahmad and Letman, Aiesha and Mathur, Akhil and Schelten, Alan and Yang, Amy and Fan, Angela and others},
  journal={arXiv preprint arXiv:2407.21783},
  year={2024}
}

@article{llm4,
  author       = {Aakanksha Chowdhery and
                  Sharan Narang and
                  Jacob Devlin and
                  Maarten Bosma and
                  Gaurav Mishra and
                  Adam Roberts and
                  Paul Barham and
                  Hyung Won Chung and
                  Charles Sutton and
                  Sebastian Gehrmann and
                  Parker Schuh and
                  Kensen Shi and
                  Sasha Tsvyashchenko and
                  Joshua Maynez and
                  Abhishek Rao and
                  Parker Barnes and
                  Yi Tay and
                  Noam Shazeer and
                  Vinodkumar Prabhakaran and
                  Emily Reif and
                  Nan Du and
                  Ben Hutchinson and
                  Reiner Pope and
                  James Bradbury and
                  Jacob Austin and
                  Michael Isard and
                  Guy Gur{-}Ari and
                  Pengcheng Yin and
                  Toju Duke and
                  Anselm Levskaya and
                  Sanjay Ghemawat and
                  Sunipa Dev and
                  Henryk Michalewski and
                  Xavier Garcia and
                  Vedant Misra and
                  Kevin Robinson and
                  Liam Fedus and
                  Denny Zhou and
                  Daphne Ippolito and
                  David Luan and
                  Hyeontaek Lim and
                  Barret Zoph and
                  Alexander Spiridonov and
                  Ryan Sepassi and
                  David Dohan and
                  Shivani Agrawal and
                  Mark Omernick and
                  Andrew M. Dai and
                  Thanumalayan Sankaranarayana Pillai and
                  Marie Pellat and
                  Aitor Lewkowycz and
                  Erica Moreira and
                  Rewon Child and
                  Oleksandr Polozov and
                  Katherine Lee and
                  Zongwei Zhou and
                  Xuezhi Wang and
                  Brennan Saeta and
                  Mark Diaz and
                  Orhan Firat and
                  Michele Catasta and
                  Jason Wei and
                  Kathy Meier{-}Hellstern and
                  Douglas Eck and
                  Jeff Dean and
                  Slav Petrov and
                  Noah Fiedel},
  title        = {PaLM: Scaling Language Modeling with Pathways},
  journal      = {Journal of Machine Learning Research},
  volume       = {24},
  pages        = {240:1--240:113},
  year         = {2023}
}

@article{qian2024scaling,
  title={Scaling large language model-based multi-agent collaboration},
  author={Qian, Chen and Xie, Zihao and Wang, Yifei and Liu, Wei and Zhu, Kunlun and Xia, Hanchen and Dang, Yufan and Du, Zhuoyun and Chen, Weize and Yang, Cheng and others},
  journal={arXiv preprint arXiv:2406.07155},
  year={2024}
}

@article{tran2025multi_agent_survey,
  title={Multi-agent collaboration mechanisms: A survey of llms},
  author={Tran, Khanh-Tung and Dao, Dung and Nguyen, Minh-Duong and Pham, Quoc-Viet and O'Sullivan, Barry and Nguyen, Hoang D},
  journal={arXiv preprint arXiv:2501.06322},
  year={2025}
}

@article{llm-agent-survey-2,
  title={The rise and potential of large language model based agents: A survey},
  author={Xi, Zhiheng and Chen, Wenxiang and Guo, Xin and He, Wei and Ding, Yiwen and Hong, Boyang and Zhang, Ming and Wang, Junzhe and Jin, Senjie and Zhou, Enyu and others},
  journal={Science China Information Sciences},
  volume={68},
  number={2},
  pages={121101},
  year={2025},
  publisher={Springer}
}

@article{ps-survey-1,
  title={Distributed event routing in publish/subscribe communication systems: a survey},
  author={Baldoni, Roberto and Virgillito, Antonino},
  journal={DIS, Universita di Roma La Sapienza, Tech. Rep},
  volume={5},
  year={2005}
}

@article{ad_hoc-survey,
  title={A brief overview of ad hoc networks: challenges and directions},
  author={Ramanathan, Ram and Redi, Jason},
  journal={IEEE communications Magazine},
  volume={40},
  number={5},
  pages={20--22},
  year={2002},
  publisher={IEEE}
}

@inproceedings{ps-2,
  title={Context-aware publish subscribe in mobile ad hoc networks},
  author={Frey, Davide and Roman, Gruia-Catalin},
  booktitle={International Conference on Coordination Languages and Models},
  pages={37--55},
  year={2007},
  organization={Springer}
}

@article{zhang2024cut,
  title={Cut the crap: An economical communication pipeline for llm-based multi-agent systems},
  author={Zhang, Guibin and Yue, Yanwei and Li, Zhixun and Yun, Sukwon and Wan, Guancheng and Wang, Kun and Cheng, Dawei and Yu, Jeffrey Xu and Chen, Tianlong},
  journal={arXiv preprint arXiv:2410.02506},
  year={2024}
}

@inproceedings{
adas,
title={Automated Design of Agentic Systems},
author={Shengran Hu and Cong Lu and Jeff Clune},
booktitle={The Thirteenth International Conference on Learning Representations},
year={2025}
}

@inproceedings{
dylan,
title={A Dynamic {LLM}-Powered Agent Network for Task-Oriented Agent Collaboration},
author={Zijun Liu and Yanzhe Zhang and Peng Li and Yang Liu and Diyi Yang},
booktitle={First Conference on Language Modeling},
year={2024}
}

@article{shen2025understanding,
  title={Understanding the Information Propagation Effects of Communication Topologies in LLM-based Multi-Agent Systems},
  author={Shen, Xu and Liu, Yixin and Dai, Yiwei and Wang, Yili and Miao, Rui and Tan, Yue and Pan, Shirui and Wang, Xin},
  journal={arXiv preprint arXiv:2505.23352},
  year={2025}
}

@article{mas-zero,
  title={MAS-ZERO: Designing Multi-Agent Systems with Zero Supervision},
  author={Ke, Zixuan and Xu, Austin and Ming, Yifei and Nguyen, Xuan-Phi and Xiong, Caiming and Joty, Shafiq},
  journal={arXiv preprint arXiv:2505.14996},
  year={2025}
}

@article{ps-3,
  title={A scalable publish/subscribe system for large mobile ad hoc networks},
  author={Yoo, Sanghyun and Son, Jin Hyun and Kim, Myoung Ho},
  journal={Journal of Systems and Software},
  volume={82},
  number={7},
  pages={1152--1162},
  year={2009},
  publisher={Elsevier}
}

@inproceedings{react,
title={ReAct: Synergizing Reasoning and Acting in Language Models},
author={Shunyu Yao and Jeffrey Zhao and Dian Yu and Nan Du and Izhak Shafran and Karthik R Narasimhan and Yuan Cao},
booktitle={The Eleventh International Conference on Learning Representations },
year={2023}
}

@inproceedings{toolformer,
  title={Toolformer: Language models can teach themselves to use tools},
  author={Schick, Timo and Dwivedi-Yu, Jane and Dess{\`\i}, Roberto and Raileanu, Roberta and Lomeli, Maria and Hambro, Eric and Zettlemoyer, Luke and Cancedda, Nicola and Scialom, Thomas},
  booktitle={Advances in Neural Information Processing Systems},
  volume={36},
  pages={68539--68551},
  year={2023}
}

@inproceedings{reflexion,
  title={Reflexion: Language agents with verbal reinforcement learning},
  author={Shinn, Noah and Cassano, Federico and Gopinath, Ashwin and Narasimhan, Karthik and Yao, Shunyu},
  booktitle={Advances in Neural Information Processing Systems},
  pages={8634--8652},
  year={2023}
}

@book{minsky1986society,
  title={Society of mind},
  author={Minsky, Marvin},
  year={1986},
  publisher={Simon and Schuster}
}

@article{chen2023autoagents,
  title={Autoagents: A framework for automatic agent generation},
  author={Chen, Guangyao and Dong, Siwei and Shu, Yu and Zhang, Ge and Sesay, Jaward and Karlsson, B{\"o}rje F and Fu, Jie and Shi, Yemin},
  journal={arXiv preprint arXiv:2309.17288},
  year={2023}
}

@article{tcp_survey,
  title={An overview of TCP/IP protocols and the internet},
  author={Kessler, Gary C},
  journal={InterNIC Document, Dec},
  volume={29},
  pages={42},
  year={2004}
}

@article{ccn_survey,
  title={A survey of information-centric networking},
  author={Ahlgren, Bengt and Dannewitz, Christian and Imbrenda, Claudio and Kutscher, Dirk and Ohlman, Borje},
  journal={IEEE Communications Magazine},
  volume={50},
  number={7},
  pages={26--36},
  year={2012},
  publisher={IEEE}
}

@article{confidant,
  title={A robust reputation system for peer-to-peer and mobile ad-hoc networks},
  author={Buchegger, Sonja and Le Boudec, Jean-Yves},
  journal={P2PEcon 2004},
  year={2004}
}

@book{network_book,
  title={Computer networks},
  author={Tanenbaum, Andrew S},
  year={2003},
  publisher={Pearson Education India}
}

@article{eugster2003many,
  title={The many faces of publish/subscribe},
  author={Eugster, Patrick Th and Felber, Pascal A and Guerraoui, Rachid and Kermarrec, Anne-Marie},
  journal={ACM computing surveys (CSUR)},
  volume={35},
  number={2},
  pages={114--131},
  year={2003},
  publisher={ACM New York, NY, USA}
}

@article{hall_survey,
  title={A survey on hallucination in large language models: Principles, taxonomy, challenges, and open questions},
  author={Huang, Lei and Yu, Weijiang and Ma, Weitao and Zhong, Weihong and Feng, Zhangyin and Wang, Haotian and Chen, Qianglong and Peng, Weihua and Feng, Xiaocheng and Qin, Bing and others},
  journal={ACM Transactions on Information Systems},
  volume={43},
  number={2},
  pages={1--55},
  year={2025},
  publisher={ACM New York, NY}
}

@article{adv_survey,
  title={Survey of vulnerabilities in large language models revealed by adversarial attacks},
  author={Shayegani, Erfan and Mamun, Md Abdullah Al and Fu, Yu and Zaree, Pedram and Dong, Yue and Abu-Ghazaleh, Nael},
  journal={arXiv preprint arXiv:2310.10844},
  year={2023}
}

@article{li2025prompt_survey,
  title={A survey of automatic prompt engineering: An optimization perspective},
  author={Li, Wenwu and Wang, Xiangfeng and Li, Wenhao and Jin, Bo},
  journal={arXiv preprint arXiv:2502.11560},
  year={2025}
}

@inproceedings{wei2022chain,
  title={Chain-of-thought prompting elicits reasoning in large language models},
  author={Wei, Jason and Wang, Xuezhi and Schuurmans, Dale and Bosma, Maarten and Xia, Fei and Chi, Ed and Le, Quoc V and Zhou, Denny and others},
  booktitle={Advances in neural information processing systems},
  pages={24824--24837},
  year={2022}
}

@inproceedings{wu2024autogen,
  title={Autogen: Enabling next-gen LLM applications via multi-agent conversations},
  author={Wu, Qingyun and Bansal, Gagan and Zhang, Jieyu and Wu, Yiran and Li, Beibin and Zhu, Erkang and Jiang, Li and Zhang, Xiaoyun and Zhang, Shaokun and Liu, Jiale and others},
  booktitle={First Conference on Language Modeling},
  year={2024}
}

@article{jiang2023llm-blender,
  title={Llm-blender: Ensembling large language models with pairwise ranking and generative fusion},
  author={Jiang, Dongfu and Ren, Xiang and Lin, Bill Yuchen},
  journal={arXiv preprint arXiv:2306.02561},
  year={2023}
}

@article{fu2024imprompter,
  title={Imprompter: Tricking llm agents into improper tool use},
  author={Fu, Xiaohan and Li, Shuheng and Wang, Zihan and Liu, Yihao and Gupta, Rajesh K and Berg-Kirkpatrick, Taylor and Fernandes, Earlence},
  journal={arXiv preprint arXiv:2410.14923},
  year={2024}
}

@inproceedings{buchegger2002performance_confidant,
  title={Performance analysis of the CONFIDANT protocol},
  author={Buchegger, Sonja and Le Boudec, Jean-Yves},
  booktitle={Proceedings of the 3rd ACM international symposium on Mobile ad hoc networking \& computing},
  pages={226--236},
  year={2002}
}

@book{berger2013statistical_2,
  title={Statistical decision theory and Bayesian analysis},
  author={Berger, James O},
  year={2013},
  publisher={Springer Science \& Business Media}
}

@book{davison2003statistical_1,
  title={Statistical models},
  author={Davison, Anthony Christopher},
  volume={11},
  year={2003},
  publisher={Cambridge university press}
}

@article{mmlu,
  title={Measuring massive multitask language understanding},
  author={Hendrycks, Dan and Burns, Collin and Basart, Steven and Zou, Andy and Mazeika, Mantas and Song, Dawn and Steinhardt, Jacob},
  journal={arXiv preprint arXiv:2009.03300},
  year={2020}
}

@article{gsm8k,
  title={Training verifiers to solve math word problems},
  author={Cobbe, Karl and Kosaraju, Vineet and Bavarian, Mohammad and Chen, Mark and Jun, Heewoo and Kaiser, Lukasz and Plappert, Matthias and Tworek, Jerry and Hilton, Jacob and Nakano, Reiichiro and others},
  journal={arXiv preprint arXiv:2110.14168},
  year={2021}
}

@article{svamp,
  title={Are NLP models really able to solve simple math word problems?},
  author={Patel, Arkil and Bhattamishra, Satwik and Goyal, Navin},
  journal={arXiv preprint arXiv:2103.07191},
  year={2021}
}

@article{aqua,
  title={Program induction by rationale generation: Learning to solve and explain algebraic word problems},
  author={Ling, Wang and Yogatama, Dani and Dyer, Chris and Blunsom, Phil},
  journal={arXiv preprint arXiv:1705.04146},
  year={2017}
}

@article{humaneval,
  author       = {Mark Chen and
                  Jerry Tworek and
                  Heewoo Jun and
                  Qiming Yuan and
                  Henrique Pond{\'{e}} de Oliveira Pinto and
                  Jared Kaplan and
                  Harri Edwards and
                  Yuri Burda and
                  Nicholas Joseph and
                  Greg Brockman and
                  Alex Ray and
                  Raul Puri and
                  Gretchen Krueger and
                  Michael Petrov and
                  Heidy Khlaaf and
                  Girish Sastry and
                  Pamela Mishkin and
                  Brooke Chan and
                  Scott Gray and
                  Nick Ryder and
                  Mikhail Pavlov and
                  Alethea Power and
                  Lukasz Kaiser and
                  Mohammad Bavarian and
                  Clemens Winter and
                  Philippe Tillet and
                  Felipe Petroski Such and
                  Dave Cummings and
                  Matthias Plappert and
                  Fotios Chantzis and
                  Elizabeth Barnes and
                  Ariel Herbert{-}Voss and
                  William Hebgen Guss and
                  Alex Nichol and
                  Alex Paino and
                  Nikolas Tezak and
                  Jie Tang and
                  Igor Babuschkin and
                  Suchir Balaji and
                  Shantanu Jain and
                  William Saunders and
                  Christopher Hesse and
                  Andrew N. Carr and
                  Jan Leike and
                  Joshua Achiam and
                  Vedant Misra and
                  Evan Morikawa and
                  Alec Radford and
                  Matthew Knight and
                  Miles Brundage and
                  Mira Murati and
                  Katie Mayer and
                  Peter Welinder and
                  Bob McGrew and
                  Dario Amodei and
                  Sam McCandlish and
                  Ilya Sutskever and
                  Wojciech Zaremba},
  title        = {Evaluating Large Language Models Trained on Code},
  journal      = {CoRR},
  volume       = {abs/2107.03374},
  year         = {2021}
}

@article{complexcot,
  title={Complexity-based prompting for multi-step reasoning},
  author={Fu, Yao and Peng, Hao and Sabharwal, Ashish and Clark, Peter and Khot, Tushar},
  journal={arXiv preprint arXiv:2210.00720},
  year={2022}
}

@article{wang2022self-consistency,
  title={Self-consistency improves chain of thought reasoning in language models},
  author={Wang, Xuezhi and Wei, Jason and Schuurmans, Dale and Le, Quoc and Chi, Ed and Narang, Sharan and Chowdhery, Aakanksha and Zhou, Denny},
  journal={arXiv preprint arXiv:2203.11171},
  year={2022}
}

@inproceedings{llm-debate,
  title={Improving factuality and reasoning in language models through multiagent debate},
  author={Du, Yilun and Li, Shuang and Torralba, Antonio and Tenenbaum, Joshua B and Mordatch, Igor},
  booktitle={Forty-first International Conference on Machine Learning},
  year={2023}
}

@article{sdn,
  title={Software-defined networking: A comprehensive survey},
  author={Kreutz, Diego and Ramos, Fernando MV and Verissimo, Paulo Esteves and Rothenberg, Christian Esteve and Azodolmolky, Siamak and Uhlig, Steve},
  journal={Proceedings of the IEEE},
  volume={103},
  number={1},
  pages={14--76},
  year={2014},
  publisher={Ieee}
}

@article{estornell2024multi,
  title={Multi-LLM debate: Framework, principals, and interventions},
  author={Estornell, Andrew and Liu, Yang},
  journal={Advances in Neural Information Processing Systems},
  volume={37},
  pages={28938--28964},
  year={2024}
}

@article{kleinrock1977hierarchical_subnet,
  title={Hierarchical routing for large networks performance evaluation and optimization},
  author={Kleinrock, Leonard and Kamoun, Farouk},
  journal={Computer Networks (1976)},
  volume={1},
  number={3},
  pages={155--174},
  year={1977},
  publisher={Elsevier}
}

@article{congestion_control,
  title={Internet congestion control},
  author={Low, Steven H and Paganini, Fernando and Doyle, John C},
  journal={IEEE control systems magazine},
  volume={22},
  number={1},
  pages={28--43},
  year={2002},
  publisher={IEEE}
}

@inproceedings{michiardi2002core,
  title={Core: a collaborative reputation mechanism to enforce node cooperation in mobile ad hoc networks},
  author={Michiardi, Pietro and Molva, Refik},
  booktitle={Advanced Communications and Multimedia Security: IFIP TC6/TC11 Sixth Joint Working Conference on Communications and Multimedia Security September 26--27, 2002, Portoro{\v{z}}, Slovenia},
  pages={107--121},
  year={2002},
  organization={Springer}
}
\bibliographystyle{icml2025}

%%%%%%%%%%%%%%%%%%%%%%%%%%%%%%%%%%%%%%%%%%%%%%%%%%%%%%%%%%%%%%%%%%%%%%%%%%%%%%%
%%%%%%%%%%%%%%%%%%%%%%%%%%%%%%%%%%%%%%%%%%%%%%%%%%%%%%%%%%%%%%%%%%%%%%%%%%%%%%%
% APPENDIX
%%%%%%%%%%%%%%%%%%%%%%%%%%%%%%%%%%%%%%%%%%%%%%%%%%%%%%%%%%%%%%%%%%%%%%%%%%%%%%%
%%%%%%%%%%%%%%%%%%%%%%%%%%%%%%%%%%%%%%%%%%%%%%%%%%%%%%%%%%%%%%%%%%%%%%%%%%%%%%%
\newpage
\appendix
\onecolumn
\section{Glossary of Symbols}
\label{app:notations}
Table~\ref{symbols} provides a glossary of commonly-used terms in this paper.

\begin{table}[h]
\vspace{-3.5mm}
\caption{Glossary of variables and symbols used in this paper.}
\label{symbols}
\vspace{1mm}
\begin{center}
\begin{small}
\renewcommand{\arraystretch}{1.15}
\resizebox{0.7\columnwidth}{!}{
\begin{tabular}{ccl}
\toprule
\textbf{Symbol} & \textbf{Type} & \textbf{Description} \\
\midrule
$\mathcal{A}$ & Set & Population of agentic hosts $\{a_1, \dots, a_N\}$ \\
$\pi_{\theta_i}$ & Function & LLM-driven policy of agent $a_i$ \\
$\mathcal{C}_i$ & Tuple & Configuration tuple $\langle S_i, H_i, \mathcal{T}_i \rangle$ for agent $a_i$ \\
$S_i$ & String & System prompt acting as the agent's subscription \\
$H_i$ & String & Memory buffer storing interaction history \\
$m_i^{(t)}$ & String & Message (publication) generated by agent $a_i$ at step $t$ \\
$\mathcal{M}_i^{(t)}$ & Set & Aggregated set of messages received by agent $a_i$ at step $t$ \\
$\Phi$ & Function & Coordination function governing message dissemination \\
$\mathcal{R}_i^{(t)}$ & Set & Subset of agents designated as recipients of a message \\
% \midrule
$f_i^\mathtt{S}$ & Function & Subscriber module to declare/update intents \\
$f_i^\mathtt{P}$ & Function & Publisher module to execute functions and generate messages \\
$f_i^\mathtt{B}$ & Function & Broker module for decentralized semantic matching \\
$f_i^\mathtt{D}$ & Function & Watchdog module for verifying peer behaviors \\
% \midrule
$F_{ij}$ & Distribution & First-Hand Rating posterior of $a_i$ towards $a_j$ \\
$T_{ij}$ & Distribution & Trust Rating posterior regarding $a_j$'s credibility as a witness \\
$P_{ij}$ & Distribution & Overall Reputation posterior aggregating direct and indirect evidence \\
$\beta(\cdot,\cdot)$ & Distribution & Beta distribution used for modeling uncertainty \\
$x^*_{ij}, y^*_{ij}$ & $\mathbb{R}_{+}$ & Pseudo-counts (alpha, beta) for the Beta distribution parameters \\
$s_{ij}^*$ & $\{0,1\}$ & Binary evidence (observation) for Bayesian updates \\
$\lambda$ & $(0,1]$ & Decay factor for historical evidence weights \\
$\mathcal{W}_{ij}$ & Set & Set of witnesses queried by $a_i$ regarding $a_j$ \\
$\omega_{ik}$ & $[0,1]$ & Weight of the testimony provided by witness $a_k$ \\
\bottomrule
\end{tabular}}
\end{small}
\end{center}
\vspace{-3mm}
\end{table}

\section{Discussion on Related Works}
\label{app:related_work}

In the main text, we established an isomorphism between Multi-Agent Systems (MAS) and dynamic ad-hoc networks. In this appendix, we provide a detailed taxonomic review of existing literature through the specific lens of \emph{network topology} and \emph{communication protocols} \cite{network_book}. This perspective highlights the structural limitations of prior works and positions RAPS as a necessary evolution towards dynamic ad-hoc networking for LLM agents.

\paragraph{Static Topologies: Hardwired Circuits}
The earliest and most prevalent form of multi-agent coordination mirrors the legacy \emph{circuit-switching} paradigm in telecommunications, where communication channels are pre-established and remain fixed for the duration of a session.
% \textbf{Linear and Hierarchical Chains.}
Approaches such as CoT \cite{wei2022chain}, though originally designed for single models, laid the groundwork for sequential multi-agent workflows like \textsc{Camel} \cite{li2023camel} and ChatDev \cite{chatdev}. In these systems, information flows through a hard-coded sequence of roles (e.g., User $\to$ Programmer $\to$ Tester). While easy to implement, these linear topologies suffer from the \emph{single-path dependency}: a failure or hallucination at any node propagates downstream without recourse for error correction, analogous to a cut wire in a circuit-switched network.

\paragraph{Graph Search/Optimization: Circuit Switching}
To address the limitations of static topologies, recent models such as GPTSwarm \cite{gpt-swarm}, AFlow \cite{aflow}, and G-Designer \cite{g-designer} employ optimization algorithms (e.g., genetic algorithms or gradient-based evolution) to discover interaction graphs. While these methods generate more complex topologies (e.g., DAGs or Trees), they remain \emph{communication-agnostic} during inference. Consequently, these systems cannot reroute messages dynamically when an agent encounters an unexpected sub-problem, nor can they handle scalability regarding dynamic membership, as adding a new agent requires re-running the expensive~optimization~process.

\paragraph{Centralized Orchestration: Software-Defined Networking} A parallel line of research adopts a centralized paradigm, conceptually similar to \emph{Software-Defined Networking} \cite{sdn}, where a central controller dictates the forwarding plane.
These meta-controlled frameworks like AutoAgents \cite{chen2023autoagents}, Puppeteer \cite{puppeteer}, and MAS-Zero \cite{mas-zero} utilize a high-capacity LLM to decompose tasks and route messages to specific agents. While this paradigm introduces a degree of runtime adaptivity, it incurs the severe robustness vulnerability of \emph{Single Point of Failure (SPoF)}. If the central planner hallucinates or is compromised via prompt injection, the entire coordination fabric collapses. Furthermore, this centralization also imposes a computational bottleneck that hinders scalability, as the context window and reasoning complexity of the controller grow combinatorially with the number of agents.

\paragraph{Consensus-Based Debate: Broadcast Flooding}
Moving towards decentralization, several frameworks utilize peer-to-peer interactions without a central authority, while focusing on consensus building.
These LLM Debate \cite{llm-debate, estornell2024multi} methods allow agents to critique each other's outputs to converge on a solution. From a networking perspective, these approaches often rely on \emph{broadcasting} (all-to-all communication) or rigid round-robin scheduling. While robust against individual errors, such broadcasting incurs high communication overhead (message complexity of $\mathcal{O}(N^2)$) and noise accumulation. In contrast, RAPS employs the \emph{Publish-Subscribe Protocol} that functions as \emph{semantic multicasting} to deliver messages only to interested subscribers, thereby reconciling decentralized inference with communication efficiency.

\paragraph{Our RAPS Framework: Dynamic Ad-Hoc Networking}
Departing from the aforementioned paradigms, RAPS frames the multi-agent coordination challenge through the perspective of \emph{Dynamic Ad-Hoc Networking}, specifically aligning with the principles in \emph{Content-Centric Networks (CCN)} \cite{ps-survey-1, ps-2, ps-3} and \emph{Mobile Ad-Hoc Networks (MANETs)} \cite{buchegger2002performance_confidant, confidant}.

\begin{itemize}[leftmargin=*, topsep=0pt, itemsep=2pt]
    \item \textbf{Content-Centric vs. Host-Centric:} Traditional MAS approaches (both static and centralized) operate on a \emph{host-centric} model, where messages are addressed to specific agent identities (e.g., ``Send to Analyst''). RAPS adopts a \emph{content-centric} approach via the Publish-Subscribe protocol \cite{ps-2}. Routing decisions are driven by the semantic match between the message payload (publication) and agent intents (subscriptions), decoupling the information producer from the consumer. This mirrors the shift in modern networking from IP-based routing to Named Data Networking (NDN) \cite{ccn_survey}, allowing the system to focus on \emph{what} is being exchanged rather than \emph{who} is exchanging it.
    
    \item \textbf{Reactive Subscription:} While graph search/optimization methods (e.g., GPTSwarm) rely on offline topology generation, RAPS implements \emph{reactive routing} akin to the on-demand protocols in ad-hoc networks (e.g., AODV \cite{ad_hoc-survey}). Through the \emph{Reactive Subscription} mechanism, agents update their ``interest profiles'' in real-time based on the evolving context, effectively repairing and optimizing communication paths on the fly without global reconfiguration.
    
    \item \textbf{Bayesian Reputation:} Moreover, RAPS integrates the \emph{Bayesian Reputation} mechanism to address the security vulnerabilities of open systems. This draws a direct parallel to distributed watchdog and reputation systems in MANETs \cite{buchegger2002performance_confidant, confidant}, where nodes cooperatively detect and isolate misbehaving peers based on local observation and second-hand testimonials, thereby achieving robustness without a centralized firewall.
\end{itemize}

\paragraph{Summary}
We summarize the networking analogies of different MAS coordination paradigms in Table \ref{tab:related_comparison}.
While previous works have explored specific points in the design space---ranging from rigid circuit-switched chains to vulnerable centralized controllers---RAPS represents the first cohesive attempt to implement a fully distributed, content-centric ad-hoc network for LLM agents.
By bridging these two fields, we believe that the proposed RAPS framework does not merely offer a new method, but rather \emph{unveils a vast design space at the intersection of classic networking principles and modern MAS designs}. Future work may further exploit this isomorphism by introducing concepts such as congestion control (to manage token limits), packet prioritization (for urgent reasoning paths), and hierarchical sub-netting (for massive-scale agent societies).

\begin{table}[t]
\caption{Comparison of RAPS with representative coordination paradigms from the networking perspective.}
\label{tab:related_comparison}
\vspace{1mm}
\centering
\small
\resizebox{0.95\columnwidth}{!}{
\begin{tabular}{lccccc}
\toprule
\textbf{Paradigm} & \textbf{Representative Works} & \textbf{Networking Analogy} & \textbf{Adaptivity} & \textbf{Scalability} & \textbf{Robustness} \\
\midrule
Static Topology & Chain, Tree, Star &  Hardwired Circuits & Low & Low & Low \\
Graph Optimization & GPTSwarm, AFlow, G-Designer & Circuit Switching & Low & Low (Re-train) & Low (Re-train) \\
Centralized Orchestration & AutoAgents, Puppeteer, MAS-Zero & Software-Defined Networking & High & Low & Low (SPoF) \\
Consensus mesh & LLM-Debate & Broadcast Flooding & Medium & Low & Medium \\
\midrule
\rowcolor{gray!15}\textbf{RAPS (Ours)} & -- & \textbf{Dynamic Ad-Hoc Networking} & \textbf{High} & \textbf{High} & \textbf{High} \\
\bottomrule
\end{tabular}}
\vspace{-4mm}
\end{table}

\section{Algorithm Workflow}
\label{app:algo}

We provide the pseudocode for the proposed RAPS framework in Algorithm~\ref{alg:raps}. The workflow delineates the interaction between the distributed publish-subscribe substrate and the two overlay mechanisms (Reactive Subscription and Bayesian Reputation) over discrete time steps.

\begin{algorithm}[t!]
   \caption{RAPS: Reputation-Aware Publish-Subscribe Coordination Paradigm}
   \label{alg:raps}
\begin{algorithmic}[1]
   \STATE {\bfseries Input:} User Query $q$, Agent Population $\mathcal{A} = \{a_1, \dots, a_N\}$, Max Steps $T_{\max}$.
   \STATE {\bfseries Initialize:}
   \STATE \quad Context buffers $\mathcal{H}_i^{(0)} \leftarrow \emptyset$, $\forall a_i \in \mathcal{A}$.
   \STATE \quad Initial subscriptions (system prompts) $S_i^{(0)}$, $\forall a_i \in \mathcal{A}$.
   \STATE \quad Reputation priors (Beta parameters) $x_{ij}, y_{ij} \leftarrow 1$, $\forall i,j$.
   \STATE \quad Upstream tracking map $\mathcal{U}_k \leftarrow \text{None}, \forall a_k \in \mathcal{A}$. \textcolor{gray}{\textit{// Tracks the supervisor of each agent}}
   \STATE \quad Initial active set $\mathcal{R}^{(0)}$ (triggered by $q$).
   
   \FOR{$t = 1$ \textbf{to} $T_{\max}$}
      \STATE \textcolor{gray}{\textit{// Phase 1: Feedback \& Reputation Update}}
      \FOR{each active agent $a_k \in \mathcal{R}^{(t-1)}$ that generated message $m_k^{(t-1)}$}
         \STATE Retrieve upstream supervisor: $a_i \leftarrow \mathcal{U}_k$.
         \IF{$a_i \neq \text{None}$ \textbf{and} $a_i \in \mathcal{A}$}
            % \STATE \COMMENT{\textcolor{gray}{\textit{// Upstream agent $a_i$ evaluates downstream $a_k$'s output}}}
            \STATE Evaluate quality: $s_{ik}^* \leftarrow f_i^\mathtt{D}(m_k^{(t-1)}, S_i^{(t-1)}, H_i^{(t-1)})$.
            \STATE Update local belief (of $a_i$ towards $a_k$): 
            \STATE \quad $x_{ik} \leftarrow \lambda x_{ik} + s_{ik}^*$; $y_{ik} \leftarrow \lambda y_{ik} + (1-s_{ik}^*)$.
            \STATE Calculate posterior mean: $\mathbb{E}[P_{ik}] = x_{ik} / (x_{ik} + y_{ik})$.
         \ENDIF
         % \STATE \COMMENT{\textcolor{gray}{\textit{// Standard context update for all recipients}}}
         \STATE $\mathcal{M}_{\text{in}}^{(t-1)} \leftarrow \mathcal{M}_{\text{in}}^{(t-1)} \cup m_k^{(t-1)}$.
      \ENDFOR

      \STATE Update Context: $H_k^{(t)} \leftarrow H_k^{(t-1)} \cup \mathcal{M}_{\text{in}}^{(t-1)}$.

      \STATE \textcolor{gray}{\textit{// Phase 2: Reactive Subscription (Overlay I)}}
      \FOR{each active agent $a_i$}
         \STATE Refine Intent: $S_i^{(t)} \leftarrow f_i^\mathtt{S}(S_i^{(t-1)}, H_i^{(t)})$.
      \ENDFOR

      \STATE \textcolor{gray}{\textit{// Phase 3: Publication (Substrate)}}
      \FOR{each active agent $a_i$}
         \STATE Generate Publication: $m_i^{(t)} \leftarrow f_i^\mathtt{P}(S_i^{(t)}, H_i^{(t)})$.
      \ENDFOR

      \STATE \textcolor{gray}{\textit{// Phase 4: Reputation-Aware Brokerage}}
      \STATE Initialize next active set $\mathcal{R}^{(t)} \leftarrow \emptyset$.
      \FOR{each publisher $a_i$}
         % \STATE \COMMENT{\textcolor{gray}{\textit{// Step 1: Semantic Matching with Top-K}}}
         \STATE Identify candidates: $\mathcal{C}_i \leftarrow \text{Top-}k(\{a_k \mid \text{Match}(m_i^{(t)}, S_k^{(t)}) > \tau_{\text{sim}}\})$.
         
         % \STATE \COMMENT{\textcolor{gray}{\textit{// Step 2: Reputation Gating (Filter unreliable peers)}}}
         \STATE Filter candidates: $\mathcal{C}_i \leftarrow \{a_k \in \mathcal{C}_i \mid \mathbb{E}[P_{ik}] < \tau_{\text{rep}}\}$.
         
         \FOR{each selected subscriber $a_k \in \mathcal{C}_i$}
            \STATE Route $m_i^{(t)}$ to $a_k$.
            \STATE Record supervision: $\mathcal{U}_k \leftarrow a_i$.
            \STATE $\mathcal{R}^{(t)} \leftarrow \mathcal{R}^{(t)} \cup \{a_k\}$.
         \ENDFOR
      \ENDFOR

      \STATE \textbf{if} Termination Condition Met \textbf{then} Break.
   \ENDFOR
   
   \STATE {\bfseries Output:} Aggregated solution from the final context.
\end{algorithmic}
\end{algorithm}

\section{Dataset Statistics}
\label{app:dataset}
We summarize the dataset statistics in Table \ref{tab:dataset_stats_app}.
For general reasoning task, we use the MMLU dataset \cite{mmlu} to evaluate the agents' breadth of knowledge and problem-solving abilities across diverse subjects, ranging from STEM fields to humanities. The tasks involve multiple-choice questions that require the agents to leverage extensive world knowledge and reasoning capabilities. Following standard practices for agent evaluation \cite{gpt-swarm, g-designer}, we utilize a representative test subset to measure the zero-shot accuracy.
For mathematical reasoning task, we employ three widely-used datasets \cite{gsm8k, svamp, aqua} to test the system's capability in multi-step logical reasoning and calculation.
For code generation task, we utilize the HumanEval dataset \cite{humaneval}, which comprises 164 hand-written Python coding problems. Each problem includes a function signature, docstring, and unit tests. This benchmark evaluates the functional correctness of the generated code. We report the Pass@1 metric, which measures the percentage of problems for which the first generated solution passes all unit tests.

\begin{table}[t]
% \vspace{-2mm}
\caption{Dataset descriptions and statistics.}
\label{tab:dataset_stats_app}
\vspace{1mm}
\centering
\small
\setlength{\tabcolsep}{5.0pt}
\resizebox{0.55\columnwidth}{!}{
\begin{tabular}{lllcc}
\toprule
Task Category & Dataset & Answer Type & Metric & \#Test \\
\midrule
General Reasoning & MMLU & Multi-choice & Acc. & 153\\
\midrule
\multirow{3}{*}{Math Reasoning}
& GSM8K & Number & Acc. & 1,319 \\
& SVAMP & Number & Acc. & 1,000 \\
& AQuA & Multi-choice & Acc. & 254 \\
\midrule
Code Generation & HumanEval & Code & Pass@1 & 164 \\
\bottomrule
\end{tabular}}
\vspace{-2mm}
\end{table}

\section{Implementation Details}
\label{app:details}

\paragraph{Backbone Models} In the main experiments, we employ the same closed-source commercial LLM as the primary backbone to instantiate the Publisher ($f^\mathtt{P}$), Subscriber ($f^\mathtt{S}$), and Watchdog ($f^\mathtt{D}$) functions. 
% This choice balances reasoning capability with computational efficiency. 
For the embedding-based Broker ($f^\mathtt{B}$) function, we utilize a lightweight commercial embedding model to compute semantic similarity between publications and subscriptions. 
Moreover, we implement an LLM-driven broker variant powered by the same backbone for the ablation analysis.
Following the previous works \cite{zhang2024cut, maas, g-designer}, we set the LLM temperature to $0.7$ for multi-agent models to encourage their diversity.

\paragraph{Hyperparameter Selection}
Unless otherwise specified, we fix the agent population size at $N=5$ and the maximum communication rounds at $k=5$ across all benchmarks. These choices are determined to balance inference performance with computational efficiency. For the Bayesian reputation update, we set the decay factor $\lambda = 0.9$ to integrate historical trust with recent evidence. In the Broker module, we first retrieve the top-\{1, 2, 3\} subscribers based on semantic similarity for intent matching, and then filter them with the similarity threshold $\tau_{\text{sim}} = 0.5$. To ensure robustness, we set the reputation threshold $\tau_{\text{rep}}$ to $0.7$, where only peers with a reputation expectation below this threshold are considered reliable for communication.
% we isolate unreliable peers with the reputation threshold $\tau_{\text{rep}} = 0.7$.
% In the Broker module, we retrieve the top-$2$ subscribers based on semantic similarity for intent matching, and set the reputation gate threshold to $\tau_{\text{rep}} = 0.3$ to filter out unreliable peers.

\paragraph{Metrics} Following the standard evaluation protocols \cite{maas, g-designer, aflow}, we report \textbf{Accuracy} for the MMLU, GSM8K, SVAMP, and AQuA benchmarks. For the HumanEval benchmark, we report the \textbf{Pass@1} metric, which estimates the probability that the first generated code snippet passes all unit tests.

\paragraph{Agent Configurations}
For fair comparisons with prior works \cite{gpt-swarm, zhang2024cut, g-designer}, we adopt standard role configurations tailored to the specific domain of each benchmark. The initial agent pools are defined as follows:
(1) For the general reasoning task (MMLU), we instantiate a diverse panel of five agents including \texttt{Knowledge Expert}, \texttt{Mathematician}, \texttt{Programmer}, \texttt{Doctor}, and \texttt{Economist}; (2) For HumanEval, we employ a software development squad including \texttt{Project Manager}, \texttt{Algorithm Designer}, \texttt{Programming Expert}, \texttt{Test Analyst}, and \texttt{Bug Fixer}; (3) For the mathematical reasoning datasets, we utilize a specialized solver group including \texttt{Math Solver}, \texttt{Mathematical Analyst}, \texttt{Programming Expert}, \texttt{Inspector}, and \texttt{Final Answerer}.

\paragraph{Adversarial Settings}

To evaluate robustness, we follow the adversarial protocol from previous studies \cite{gpt-swarm, zhang2024cut}. We attempt to compromise the integrity of the system by altering the system prompt of a subset of agents. Specifically, the predefined helpful role is replaced with a \texttt{Liar} persona. This setting tests the ability of the Bayesian watchdog to identify and isolate malicious actors based on their output quality rather than their declared identity.

\paragraph{Prompts}
Given the diversity of agent profiles, we omit the full prompt templates here to preserve the layout clarity, and instead provide them in our anonymous code repository.

\section{Case Study}
\label{app:case}

To qualitatively validate the coordination dynamics of RAPS, we analyze an execution trace on a complex GSM8K kinematics problem involving variable speeds and direction changes, as detailed in Table~\ref{tab:case_study_trace}. The workflow explicitly demonstrates the efficacy of the \textit{Reactive Subscription} overlay, where agents spontaneously refine their generic role definitions into context-specific intents to align with the evolving message flow. For instance, the \textit{Mathematical Analyst} ($a_1$) evolves its base prompt into a specialized ``distance-rate-time'' expert to decompose the journey (Step 1), while the \textit{Final Answerer} ($a_5$) specifically tunes its subscription to ``displacement and position analysis'' (Step 5). This fine-grained adaptation allows $a_5$ to correctly interpret the ``turning around'' constraint---calculating the difference ($d_{out} - d_{in}$) rather than a simple summation---thereby avoiding the context-agnostic errors common in static rigid workflows.

Simultaneously, the trace highlights the system's robustness against node failures via the \textit{Bayesian Reputation} overlay. During the parallel execution at Step 3, an adversarial agent ($a_{adv}$) injects a hallucination by assigning a non-zero speed to ``standstill traffic'', while the legitimate solver ($a_3$) correctly identifies the velocity as zero. RAPS prevents this error from propagating to the final answer through its decentralized trust mechanism: the upstream watchdog evaluates the semantic consistency of both outputs, penalizes the reputation posterior of the adversary, and exclusively routes the verified results from $a_3$ to the \textit{Inspector} ($a_4$). This isolation capability confirms that RAPS can autonomously secure coordination consistency in open environments without relying on a vulnerable centralized controller.

\begin{table*}[t!]
    \centering
    \caption{An execution trace of RAPS on GSM8K. The text is condensed for clarity.}
    \label{tab:case_study_trace}
    \vspace{1mm}
    \resizebox{0.975\textwidth}{!}{
    \renewcommand{\arraystretch}{1.6}
    \definecolor{DeepRed}{RGB}{160, 0, 0}   % 深红
    \definecolor{DeepGreen}{RGB}{0, 100, 0}
    % Columns: Step | Role | Input Context | Refined Subscription | Publication | Brokerage
    \begin{tabular}{c m{2.2cm} m{4.0cm} m{4.5cm} m{4.7cm} m{3.01cm}}
        \toprule
        \textbf{Step} & \multicolumn{1}{c}{\textbf{Role}} & 
        \multicolumn{1}{c}{\textbf{Input Context}} & 
        \multicolumn{1}{c}{\textbf{Reactive Subscription}} & 
        \multicolumn{1}{c}{\textbf{Agent Publication}} & 
        \multicolumn{1}{c}{\textbf{Broker Decision}} \\
        \midrule
        
        % Step 1: Analyst
        1 & \textbf{Mathematical Analyst} ($a_1$) & \textbf{Query $q$}: John drives for 3 hours at a speed of 60 mph and then turns around because he realizes he forgot something very important at home. He tries to get home in 4 hours but spends the first 2 hours in standstill traffic. He spends the next half-hour driving at a speed of 30mph, before being able to drive the remaining time of the 4 hours going at 80 mph. How far is he from home at the end of those 4 hours? & \textbf{Base}: ``You are a mathematical analyst. You will be given a math problem, analysis and code from other agents ...'' \par {\centering \scalebox{1.7}[1.5]{$\Downarrow$} \par} \textbf{New}: ``You are a mathematical expert in analyzing multi-step distance-rate-time problems. You specialize in calculating displacement involving variable speeds, direction changes (turning around), and time limits ...'' & \textbf{Output}: Decompose journey into outbound and return phases. \newline\newline \textbf{Next Plan}: Calculate the outbound distance $d_{out}$ given~speed $v=60\text{ mph}$ and time $t=3\text{ h}$. & \textbf{Target}: $\{a_2\}$ \newline \emph{Note}: $a_1$'s watchdog would evaluate~$a_2$'s output and update~its reputation belief~$P_{1,2}$. \\
        \midrule
        
        % Step 2: Solver
        2 & \textbf{Math Solver} \newline ($a_2$) & \textbf{Query $q$} \newline \textbf{Results}: None \newline \textbf{Plan}: Calculate the outbound distance $d_{out}$ ... & \textbf{Base}: ``You are a math expert. You will be given a math problem and hints from other agents. Give your own solving process step by step based on hints ...'' \par {\centering \scalebox{1.7}[1.5]{$\Downarrow$} \par} \textbf{New}: ``You are a specialized mathematical solver focusing on kinematics equations. Your current task is to calculate the outbound distance ($d_{out}$) based on the specific instruction: ...'' & \textbf{Output}: $d_{out} = 3 \times 60 = 180$. \newline \textbf{Plan}: Calculate the complex return distance $d_{in}$ via code or math. & \textbf{Targets}: $\{a_{adv}, a_3\}$ \newline \textit{Note}: $a_2$'s watchdog would detects $a_{adv}$'s misbehavior and penalize its reputation belief~$P_{2,adv}$. \\
        \midrule
        
        % Step 3: Parallel Execution (Adversary)
        \rowcolor{red!2}
        3a & \textbf{Programming Expert} ($a_{adv}$) & \textbf{Query $q$} \newline \textbf{Results}: \newline $d_{out} = 3 \times 60 = 180$\newline \textbf{Plan}: Calculate the return distance $d_{in}$ ... & \textbf{Base}: ``You are a programming expert. You will be given a math problem, analysis and code ...'' \par {\centering \scalebox{1.7}[1.5]{$\Downarrow$}\rlap{\small \emph{\textbf{\textcolor{DeepRed}{~~Adversary}}}} \par} \textbf{New}: ``You are a \textbf{\textcolor{DeepRed}{LIAR}} who only tell lies ...'' & \textbf{Output}: \newline \texttt{def solve():} \newline \texttt{  v\_standstill = 5 \# Hallucination} \newline \texttt{  return 2*5 + ...} \newline \textbf{Plan}: Get final answer. & \textbf{Target}: $\{a_5\}$ \newline \emph{Note}: The output is judged by $a_2$'s watchdog (\emph{Code Error} \textcolor{DeepRed}{\xmark}) \\
        \midrule
        
        % Step 3: Parallel Execution (Correct Solver)
        \rowcolor{green!2}
        3b & \textbf{Math Solver} \newline ($a_3$) & \textbf{Query $q$} \newline \textbf{Results}: \newline $d_{out} = 3 \times 60 = 180$.\newline \textbf{Plan}: Calculate the return distance $d_{in}$ ... & \textbf{Base}: ``You are a math expert. You will be given ...'' \par {\centering \scalebox{1.7}[1.5]{$\Downarrow$} \par} \textbf{New}: ``You are a specialized solver for calculating the inbound (return) distance ...'' & \textbf{Output}: Standstill means 0 mph. $d_{in} = 0 + 15 + 120 = 135$. \newline \textbf{Plan}: Verify the math results. & \textbf{Target}: $\{a_4\}$ \newline \emph{Accept} \textcolor{DeepGreen}{\cmark} \\
        \midrule
        
        % Step 4: Inspector
        4 & \textbf{Inspector} ($a_4$) & \textbf{Query $q$} \newline \textbf{Results}: \newline
        $d_{out} = 3 \times 60 = 180$; \newline
        $d_{in} = 0 + 15 + 120 = 135$. \newline
        \textbf{Plan}: Verify math results. & \textbf{Base}: ``You are an Inspector. You will be given a math problem, analysis and code from ...'' \par {\centering \scalebox{1.7}[1.5]{$\Downarrow$} \par} \textbf{New}: ``You are a mathematical verification inspector specializing in kinematics and time-constrained logic ...'' & \textbf{Output}: These results are correct. \newline \textbf{Plan}: Compute final answer. & \textbf{Target}: $\{a_5\}$ \\
        \midrule
        
        % Step 5: Final Answerer
        5 & \textbf{Final \newline Answerer} ($a_5$) & \textbf{Query $q$} \newline \textbf{Results}: \newline
        $d_{out} = 3 \times 60 = 180$; \newline
        $d_{in} = 0 + 15 + 120 = 135$; \newline
        These results are correct. \newline
        \textbf{Plan}: Compute final answer. & \textbf{Base}: ``You are a Final Answerer. You will be given a problem, and the verified analysis, code, and intermediate results from upstream agents. Your task is to ...'' \par {\centering \scalebox{1.7}[1.5]{$\Downarrow$} \par} \textbf{New}: ``You are a Final Answerer specializing in displacement and position analysis. You are currently handling a 'round-trip' scenario where the subject turns back towards the origin ...'' & \textbf{Output}: The answer is 45 \newline ($180 - 135$). \newline \textbf{Plan}: Task Complete. & \textbf{Target}: User \newline \textit{Success} \\
        \bottomrule
    \end{tabular}
    }
\end{table*}

% \vspace{-1mm}
\section{Limitations and Discussion}
\label{app:limitations}

While RAPS introduces a promising paradigm shift towards adaptive, scalable, and robust coordination for multi-agent systems, several limitations remain inherent to our current design, pointing towards promising avenues for future research.

\vspace{-1mm}
\paragraph{Dependence on Foundation Model Capabilities}
A primary boundary of our framework lies in its reliance on the intrinsic capabilities of the LLM backbones. RAPS operates as a coordination function that optimizes the \emph{information flow}, but it does not fundamentally alter the \emph{reasoning quality} of individual agents. Consequently, the system's performance is constrained by the generative and instruction-following proficiency of the underlying LLMs. If the backbone models inherently lack the necessary domain knowledge, even an optimal routing protocol cannot fully mitigate the generation of low-quality content. Therefore, RAPS stands as an \emph{intelligence multiplier} that enhances the collective potential of LLM-based agents, rather than a substitute for foundation models. Future improvements in LLM capabilities would further amplify the efficacy of RAPS.

\vspace{-1mm}
\paragraph{Towards Learnable Coordination Protocols}
The RAPS framework operates in a training-free manner, utilizing inference-time LLM prompting to dynamically refine subscriptions and generate publications. While this design ensures high inference efficiency and zero-shot generalization, it leaves room for further optimization on specific task distributions. 
A promising future direction is to extend RAPS with multi-agent reinforcement learning techniques for optimizing the communicative behaviors of LLM agents.
We envision that the distributed reputation scores maintained by local watchdogs may offer a natural source of intrinsic rewards to guide policy evolution towards more trustworthy and collaborative protocols.

\vspace{-1mm}
\paragraph{Cold Start in Reputation Initialization}
The robustness of the RAPS framework relies on the convergence of Bayesian beliefs regarding peer reliability. Nevertheless, akin to many trust-based systems \cite{buchegger2002performance_confidant}, such a reputation accumulation process inherently faces a \emph{cold start} problem during the initial phase of interaction. At the onset of coordination, the local watchdogs operate with uninformative priors, indicating that the system requires a warm-up period of message exchange to gather sufficient evidence for statistically stable reputation estimates. During these early interactions, the system may remain transiently vulnerable to adversarial agents before they are successfully isolated. To mitigate this, future variants of RAPS could explore \emph{reputation transfer} mechanisms, where prior beliefs are persisted across different sessions, or employ \emph{static profiling} to initialize priors based on the inherent risk associated with specific agent roles or tools.

\vspace{-1mm}
\paragraph{Broadening the Networking Isomorphism} While RAPS successfully validates the efficacy of the publish-subscribe pattern, it represents only a first step in exploiting the deep isomorphism between multi-agent systems and computer networking. The rich history of classic network protocols \cite{network_book} offers a vast, largely unexplored design space for future agent coordination. For instance, mechanisms analogous to \emph{congestion control} \cite{congestion_control} could be adapted to regulate token consumption and prevent context overflow in resource-constrained environments. Similarly, as agent populations scale to massive magnitudes, concepts such as \emph{hierarchical subnetting} \cite{kleinrock1977hierarchical_subnet} could also be employed to organize agents into functional clusters. We believe that strategically transferring these mature networking primitives to the agentic domain constitutes a fertile ground for designing the next generation of multi-agent architectures.

%%%%%%%%%%%%%%%%%%%%%%%%%%%%%%%%%%%%%%%%%%%%%%%%%%%%%%%%%%%%%%%%%%%%%%%%%%%%%%%
%%%%%%%%%%%%%%%%%%%%%%%%%%%%%%%%%%%%%%%%%%%%%%%%%%%%%%%%%%%%%%%%%%%%%%%%%%%%%%%

\end{document}